\documentclass[a4paper,fleqn]{cas-dc}

\usepackage{natbib}
\usepackage{amssymb}
\usepackage{amsmath}
\usepackage{lipsum}

\newcommand{\bigO}{\mathcal{O}}
\usepackage{hyperref}
\usepackage{nomencl}
\usepackage{graphicx} 
\usepackage{enumitem}
\usepackage{multirow}
\usepackage{pifont}
\usepackage{gensymb}
\usepackage{mathtools}
\usepackage{tikz}
\usetikzlibrary{arrows.meta, spy, calc, arrows, patterns, patterns.meta}
\usepackage{adjustbox}
\usepackage{graphicx}

\begin{document}

\shorttitle{SPARC: Automatic 3D$+$time fetal cardiac MRI reconstruction}    

\shortauthors{A. Boutillon \textit{et al.}}  

\title[mode=title]{SPARC: \textbf{S}lice-to-volume \textbf{P}ipeline for \textbf{A}utomated \textbf{R}econstruction of gated 3D$+$time fetal \textbf{C}ardiac MRI} 

\author[label1,label2]{Arnaud Boutillon}\cormark[1]\fnmark[1]
\author[label1,label3]{Naomi Clarke}\fnref{fn1}
\author[label3]{Tomas Woodgate}
\author[label2,label4]{Daniel West}
\author[label2]{Alina Schneider}
\author[label2]{Rachael Franklin}
\author[label2,label4]{Anthony Price}
\author[label2,label3]{Jo Hajnal}
\author[label3]{Kuberan Pushparajah}
\author[label3]{David Lloyd}
\author[label1,label3]{Maria Deprez}

\affiliation[label1]{
organization={Biomedical Computing Research Department, School of Biomedical Engineering and Imaging Sciences, King’s College London},
city={London},
country={United Kingdom}}
\affiliation[label2]{
organization={Imaging Physics and Engineering Research Department, School of Biomedical Engineering and Imaging Sciences, King’s College London},
city={London},
country={United Kingdom}}
\affiliation[label3]{
organization={Early Life Imaging Research Department, School of Biomedical Engineering and Imaging Sciences, King’s College London},
city={London},
country={United Kingdom}}
\affiliation[label4]{
organization={Medical Physics and Clinical Engineering, Guys and St Thomas’ NHS Foundation Trust, London, United Kingdom},
city={London},
country={United Kingdom}}

\fntext[fn1]{These authors contributed equally to this work.}
\cortext[cor1]{Corresponding author: \href{mailto:arnaud.boutillon@kcl.ac.uk}{\nolinkurl{arnaud.boutillon@kcl.ac.uk}} (A. Boutillon)}

\begin{abstract} 
Fetal cardiac MRI (fCMR) provides valuable diagnostic information complementary to echocardiography, particularly for complex congenital heart disease (CHD). Dynamic cine imaging captures cardiac motion essential for assessment of cardiac function; however, the reconstruction of 3D$+$time cine volumes from 2D$+$time acquired slices remains challenging due to unpredictable fetal motion and the absence of automated and robust processing tools suitable for clinical deployment. We present the \textbf{SPARC} pipeline (\textbf{S}lice-to-volume \textbf{P}ipeline for \textbf{A}utomated \textbf{R}econstruction of gated 3D$+$time fetal \textbf{C}ardiac MRI) which combines physics-informed slice-to-volume reconstruction (SVR) of Doppler ultrasound (DUS) gated stacks of slices, assisted by deep learning (DL) models for thoracic segmentation and anatomical reorientation. The proposed SVR algorithm achieves a tenfold reduction in reconstruction time relative to existing frame-wise approaches ($4.8 \pm 1.0$ vs $49.0 \pm 14.1$ min, $p < 0.0001$) while improving the  reconstruction quality. Thoracic segmentation performance using ensemble aggregation exceeded inter-rater agreement (Dice $84.7 \pm 3.9\%$ vs $81.4 \pm 7.7\%$, $p<0.05$), while anatomical reorientation achieved a success rate of $90.1\%$. End-to-end evaluation on a large held-out clinical cohort ($n = 121$) demonstrated fully automatic processing in $82.6\%$ of cases with a mean runtime of $7.1 \pm 1.3$ min, compatible with clinical deployment. The complete \textbf{SPARC} pipeline is publicly available as a Docker container \href{https://hub.docker.com/r/aboutill/sparc}{https://hub.docker.com/r/aboutill/sparc} and is currently deployed at our institution as a clinical research tool.
\end{abstract}


\maketitle

\section{Introduction}

Congenital heart disease (CHD) affects approximately $1\%$ of births worldwide \citep{liu_global_2019}, and prenatal detection is associated with improved postnatal outcomes \citep{gardiner_prenatal_2014, holland_prenatal_2015, lloyd_three-dimensional_2019}. Two-dimensional (2D) echocardiography remains the primary modality for prenatal cardiac screening due to its widespread availability and ease of use. However, there is a longstanding need for complementary three-dimensional (3D) visualisation of fetal cardiac anatomy due to the inherent constraints of 2D imaging \citep{lloyd_exploration_2016, lloyd_three-dimensional_2019}. Magnetic resonance imaging (MRI) has emerged as a valuable adjunct to fetal echocardiography \citep{roy_fetal_2019, dong_fetal_2020, mamalis_evolution_2022, lim_fetal_2024, vollbrecht_fetal_2024, woodgate_time-resolved_2024}, with demonstrated utility in predicting the need for early postnatal intervention \citep{lloyd_analysis_2021} and good agreement with both echocardiographic and postnatal findings \citep{moscatelli_cardiovascular_2023}.

The fetal heart is a small, complex, and dynamic organ that presents unique imaging challenges \citep{van_amerom_fetal_2018, woodgate_time-resolved_2024}. Fetal echocardiography is limited by acoustic window quality, which deteriorates towards the end of pregnancy, as well as maternal obesity and unfavourable fetal positions \citep{moscatelli_cardiovascular_2023}. Fetal cardiac magnetic resonance (fCMR) faces distinct but complementary challenges, including maternal respiratory motion, fetal motion \citep{votino_magnetic_2012}, the absence of an electrocardiogram (ECG) gating signal, and maternal discomfort \citep{vollbrecht_fetal_2024}. Maternal breath-holds and fetal sedation can improve image quality but increase discomfort and anxiety \citep{van_amerom_fetal_2018}. 

In the non-fetal context, dynamic cardiac MRI is well established using ECG gating to synchronise acquisition with the cardiac cycle and reconstruct 3D$+$time cine volumes. ECG cannot be applied in the fetal context, necessitating alternative synchronisation strategies. For example, balanced steady-state free precession (bSSFP) cine sequences have enabled recognition of basic cardiovascular anatomy but remain susceptible to fetal motion \citep{geiger_feasibility_2023}. Self-gating techniques, including metric optimised gating (MOG) and other image-based approaches \citep{roy_dynamic_2013, haris_self-gated_2017, van_amerom_fetal_2018}, address the absence of ECG but introduce additional post-processing complexity and can perform suboptimally under maternal free-breathing conditions \citep{haris_freebreathing_2020}.

\begin{figure*}[t]
\centering
\begin{tikzpicture}[every node/.style={inner sep=0,outer sep=0}]

\node at (7.5,4.5) {Fetal cardiac cycles with acquisition triggers};

\draw[color=black, domain=0:7/3, variable=\x, samples=1000] plot ({\x}, {1/2*cos(6*pi*\x/7 r) + 3.45});
\draw[color=black, domain=7/3:7/3+7/3/0.9, variable=\x, samples=1000] plot ({\x}, {1/2*cos(0.9*6*pi*(\x-7/3)/7 r) + 3.45});
\draw[color=black, domain=7/3+7/3/0.9:7/3+7/3/0.9+7/3/1.1, variable=\x, samples=1000] plot ({\x}, {1/2*cos(1.1*6*pi*(\x-7/3-7/3/0.9)/7 r) + 3.45});
\draw[color=gray, domain=8:8+7/3/1.15, variable=\x, samples=1000] plot ({\x}, {1/2*cos((1.15*6*pi*(\x-1)/7-1.15*6*pi) r) +3.45});
\draw[color=gray, domain=8+7/3/1.15:8+7/3/1.15+7/3/1.05, variable=\x, samples=1000] plot ({\x}, {1/2*cos((1.05*6*pi*(\x-1-7/3/1.15)/7-1.05*6*pi) r) +3.45});
\draw[color=gray, domain=8+7/3/1.15+7/3/1.05:8+7/3/1.15+7/3/1.05+7/3/1.25, variable=\x, samples=1000] plot ({\x}, {1/2*cos((1.25*6*pi*(\x-1-7/3/1.15-7/3/1.05)/7-1.25*6*pi) r) +3.45});

\draw[<->]  (0,2.85) -- (7/3,2.85);
\node at (7/6,2.5) {$\text{RR}_{\text{Inst}}$};
\draw[<->]  (39/15,2.85) -- (45.5/15,2.85);
\node at (42.25/15,2.5) {TR};
\draw[<->]  (8,2.85) -- (8+7/3/1.15+7/3/1.05+7/3/1.25,2.85);
\node at (8+7/3/1.15/2+7/3/1.05/2+7/3/1.25/2,2.5) {$(k+1)$-th slice acquisition};

\node at (7.5,3.45) {$\dots$};

\draw[line width=0.25mm, |-|] (0,4.1) -- (0,3.8);
\draw[line width=0.25mm, |-|] (7/3,4.1) -- (7/3,3.8);
\draw[line width=0.25mm, |-|] (7/3+7/3/0.9,4.1) -- (7/3+7/3/0.9,3.8);
\draw[line width=0.25mm, |-|] (7/3+7/3/0.9+7/3/1.1,4.1) -- (7/3+7/3/0.9+7/3/1.1,3.8);
\draw[line width=0.25mm, |-|] (8,4.1) -- (8,3.8);
\draw[line width=0.25mm, |-|] (8+7/3/1.15,4.1) -- (8+7/3/1.15,3.8);
\draw[line width=0.25mm, |-|] (8+7/3/1.15+7/3/1.05,4.1) -- (8+7/3/1.15+7/3/1.05,3.8);
\draw[line width=0.25mm, |-|] (8+7/3/1.15+7/3/1.05+7/3/1.25,4.1) -- (8+7/3/1.15+7/3/1.05+7/3/1.25,3.8);

\draw[color=blue!50!magenta, line width=0.5mm] (0,4.05) -- (0,3.85);
\draw[color=blue!50!magenta, line width=0.5mm] (6.5/15,3.55) -- (6.5/15,3.75);
\draw[color=blue!50!magenta, line width=0.5mm] (13/15,3.2) -- (13/15,3.0);
\draw[color=blue!50!magenta, line width=0.5mm] (19.5/15,3.08) -- (19.5/15,2.88);
\draw[color=blue!50!magenta, line width=0.5mm] (26/15,3.33) -- (26/15,3.53);
\draw[color=blue!50!magenta, line width=0.5mm] (32.5/15,3.8) -- (32.5/15,4.0);

\draw[color=red!50!yellow, line width=0.5mm] (39/15,3.75) -- (39/15,3.95);
\draw[color=red!50!yellow, line width=0.5mm] (45.5/15,3.29) -- (45.5/15,3.49);
\draw[color=red!50!yellow, line width=0.5mm] (52/15,2.89) -- (52/15,3.09);
\draw[color=red!50!yellow, line width=0.5mm] (58.5/15,3.15) -- (58.5/15,2.95);
\draw[color=red!50!yellow, line width=0.5mm] (65/15,3.42) -- (65/15,3.62);
\draw[color=red!50!yellow, line width=0.5mm] (72.5/15,3.84) -- (72.5/15,4.04);

\draw[color=Turquoise, line width=0.5mm] (79/15,3.62) -- (79/15,3.82);
\draw[color=Turquoise, line width=0.5mm] (85.5/15,3.02) -- (85.5/15,3.22);
\draw[color=Turquoise, line width=0.5mm] (92/15,3.1) -- (92/15,2.9);
\draw[color=Turquoise, line width=0.5mm] (98.5/15,3.43) -- (98.5/15,3.63);
\draw[color=Turquoise, line width=0.5mm] (105/15,3.85) -- (105/15,4.05);

\draw[->]  (3.5,2.2) -- (3.5,1.2);
\node[anchor=west] at (3.65,1.85) {Temp. interp.};
\node[anchor=west] at (3.65,1.5) {in $k$-space};

\draw (0,0) -- (0,1) -- (1,1) -- (1,0) -- cycle;
\draw[color=blue!50!magenta] (0.1,.9) -- (.9,.9);
\draw[color=blue!50!magenta] (0.1,.8) -- (.9,.8);
\draw[color=red!50!yellow] (0.1,.55) -- (.9,.55);
\draw[color=red!50!yellow] (0.1,.45) -- (.9,.45);
\draw[color=Turquoise] (0.1,.1) -- (.9,.1);
\draw[color=Turquoise] (0.1,.2) -- (.9,.2);
\draw[->]  (0,-0.15) -- (1,-0.15);
\draw[->]  (-0.15,0) -- (-0.15,1);
\node at (0.5,-.4) {$k_x$};
\node at (-.4,0.5) {$k_y$};

\draw (1.5,0) -- (1.5,1) -- (2.5,1) -- (2.5,0) -- cycle;
\draw[color=blue!50!magenta] (1.6,.9) -- (2.4,.9);
\draw[color=blue!50!magenta] (1.6,.8) -- (2.4,.8);
\draw[color=red!50!yellow] (1.6,.55) -- (2.4,.55);
\draw[color=red!50!yellow] (1.6,.45) -- (2.4,.45);
\draw[color=Turquoise] (1.6,.1) -- (2.4,.1);
\draw[color=Turquoise] (1.6,.2) -- (2.4,.2);

\draw (3,0) -- (3,1) -- (4,1) -- (4,0) -- cycle;
\draw[color=blue!50!magenta] (3.1,.9) -- (3.9,.9);
\draw[color=blue!50!magenta] (3.1,.8) -- (3.9,.8);
\draw[color=red!50!yellow] (3.1,.55) -- (3.9,.55);
\draw[color=red!50!yellow] (3.1,.45) -- (3.9,.45);
\draw[color=Turquoise] (3.1,.1) -- (3.9,.1);
\draw[color=Turquoise] (3.1,.2) -- (3.9,.2);

\draw (4.5,0) -- (4.5,1) -- (5.5,1) -- (5.5,0) -- cycle;
\draw[color=blue!50!magenta] (4.6,.9) -- (5.4,.9);
\draw[color=blue!50!magenta] (4.6,.8) -- (5.4,.8);
\draw[color=red!50!yellow] (4.6,.55) -- (5.4,.55);
\draw[color=red!50!yellow] (4.6,.45) -- (5.4,.45);
\draw[color=Turquoise] (4.6,.1) -- (5.4,.1);
\draw[color=Turquoise] (4.6,.2) -- (5.4,.2);

\draw (6,0) -- (6,1) -- (7,1) -- (7,0) -- cycle;
\draw[color=blue!50!magenta] (6.1,.9) -- (6.9,.9);
\draw[color=blue!50!magenta] (6.1,.8) -- (6.9,.8);
\draw[color=red!50!yellow] (6.1,.55) -- (6.9,.55);
\draw[color=red!50!yellow] (6.1,.45) -- (6.9,.45);
\draw[color=Turquoise] (6.1,.1) -- (6.9,.1);
\draw[color=Turquoise] (6.1,.2) -- (6.9,.2);

\draw[->]  (0,-.9) -- (7,-.9);
\draw  (0.5,-.75) -- (0.5,-1.05);
\draw  (2,-.75) -- (2,-1.05);
\draw  (3.5,-.75) -- (3.5,-1.05);
\draw  (5,-.75) -- (5,-1.05);
\draw  (6.5,-.75) -- (6.5,-1.05);
\node at (3.5,-1.35) {$T$ interpolated cardiac phases};

\draw[->] (7.7,0.5) -- (9.1,0.5);
\node at (8.4,0.25) {2D Fourier};
\node at (8.4,-.1) {transform};

\draw[fill=cyan] (10.2,0.2) -- (10.2,1.2) -- (11.2,1.2) -- (11.2,0.2) -- cycle;
\node[inner sep=0pt] at (10.7,0.7) 
{\includegraphics[width=.055\textwidth,angle=-90]{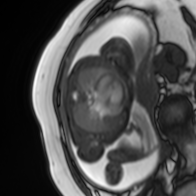}};
\draw[fill=cyan] (10.1,0.1) -- (10.1,1.1) -- (11.1,1.1) -- (11.1,0.1) -- cycle;
\node[inner sep=0pt] at (10.6,0.6) 
{\includegraphics[width=.055\textwidth,angle=-90]{acquisition/slice_ax.png}};
\draw[fill=cyan] (10,0) -- (10,1) -- (11,1) -- (11,0) -- cycle;
\node[inner sep=0pt] at (10.5,0.5) 
{\includegraphics[width=.055\textwidth,angle=-90]{acquisition/slice_ax.png}};
\draw[fill=cyan] (9.9,-0.1) -- (9.9,.9) -- (10.9,.9) -- (10.9,-0.1) -- cycle;
\node[inner sep=0pt] at (10.4,0.4) 
{\includegraphics[width=.055\textwidth,angle=-90]{acquisition/slice_ax.png}};
\draw[fill=cyan] (9.8,-0.2) -- (9.8,.8) -- (10.8,.8) -- (10.8,-0.2) -- cycle;
\node[inner sep=0pt] at (10.3,0.3) 
{\includegraphics[width=.055\textwidth,angle=-90]{acquisition/slice_ax.png}};
\draw[->]  (11,-0.2) -- (11.4,0.2);
\draw[->]  (9.8,-0.35) -- (10.8,-0.35);
\draw[->]  (9.65,-.2) -- (9.65,0.8);
\node at (10.3,-.55) {$x$};
\node at (9.45,.3) {$y$};
\node at (11.3,-0.1) {$t$};

\node[anchor=west] at (11.45,.85) {Doppler-gated};
\node[anchor=west] at (11.45,.5) {2D$+$time slice};

\end{tikzpicture}
\caption{Doppler-gated acquisition of a single 2D$+$time slice. The bSSFP sequence acquires one $k$-space segment per $\text{TR}$, with the number of acquired cardiac phases per heartbeat given by $T_{\text{acq}} = \lfloor \text{RR}_{\text{Inst}} / \text{TR} \rfloor$, where $\text{RR}_{\text{Inst}}$ is the instantaneous RR interval. Since $\text{RR}_{\text{Inst}}$ varies across heartbeats, $T_{\text{acq}}$ is not constant. Temporal interpolation of each $k$-space line to $T$ uniformly spaced cardiac phases, followed by a 2D Fourier transform, yields the reconstructed Doppler-gated 2D$+$time slice.}
\label{fig:doppler_acquisition}
\end{figure*}

The recent introduction of an MR-compatible Doppler ultrasound (DUS) gating device \citep{kording_evaluation_2018, kording_dynamic_2018}, that provides a cardiac gating signal analogous to the ECG (Figure~\ref{fig:doppler_acquisition}), enables dynamic cine acquisition of the fetal heart. DUS-gated fCMR has been shown to provide cardiac output measurements within physiologically plausible ranges and additional diagnostic value over echocardiography alone \citep{ryd_utility_2021, vollbrecht_fetal_2023, minocha_reference_2024}. Existing studies combining DUS-gating with a breath-holding to acquire 2D+time standard views demonstrated clinical feasibility but remain limited by motion \citep{ryd_utility_2021, vollbrecht_fetal_2023}. The feasibility of free-breathing 2D$+$time DUS-gated radial acquisition  with within-slice Fourier domain motion compensation has been demonstrated by \citet{haris_freebreathing_2020}. However, the extension of DUS-gated fCMR to motion-compensated 3D$+$time imaging has not yet been proposed.

In this work, we aim to reconstruct motion-compensated 3D$+$time fCMR from stacks of overlapping DUS-gated 2D$+$time slices acquired in multiple orientations in the presence of maternal respiration and fetal motion. Slice-to-volume reconstruction (SVR) methods were originally developed for 3D fetal brain reconstruction \citep{rousseau_novel_2005, kuklisova-murgasova_reconstruction_2012, gholipour_motion-corrected_2019} and applied to reconstruct static 3D fetal heart \citep{lloyd_three-dimensional_2019, uus_automated_2022}. SVR has been extended for 3D$+$time fetal cardiac MRI \citep{van_amerom_fetal_2019}, to reconstruct a cine volume spanning a single cardiac cycle from free-running, un-gated acquisitions by retrospectively estimating and synchronising the cardiac phase of each image frame using image-derived self-gating. While this approach demonstrated the feasibility of 3D$+$time fetal cardiac cine reconstruction, it required long processing times and substantial manual input including segmentation of the chest and heart and reorientation to canonical coordinate system, limiting its applicability in clinical practice. 

Deep learning has recently been applied to fCMR to improve image quality and reduce acquisition time. The work of \citet{vollbrecht_deep_2024} demonstrated that deep learning denoising and super-resolution applied to Doppler-gated fCMR produced reconstructions of higher diagnostic confidence. However, the absence of motion mitigation meant that fetal movement remained a significant source of image artefacts. \citet{berggren_superresolution_2022} proposed an alternative strategy to reduce motion sensitivity by shortening scan times, although, re-acquisition or additional post-processing was still required when fetal movement occurred. These approaches all underscore the importance of motion correction.

\citet{uus_automated_2022} proposed an automated SVR pipeline for static 3D fCMR in which deep learning models provided automatic region of interest segmentation and landmark detection to recover large inter-stack displacements. Recent work has also formulated SVR as an optimisation problem using implicit neural representations (INR) \citep{sitzmann_implicit_2020}, offering a promising alternative paradigm \citep{xu_nesvor_2023}. However, such approaches have not yet been extended to fetal cardiac 3D$+$time MR imaging.

\subsection{Contributions}

\begin{figure*}[t]
\centering
\begin{tikzpicture}[every node/.style={inner sep=0,outer sep=0}]

\draw[fill=cyan] (0.2,-3.8) -- (0.2,-4.8) -- (1.2,-4.8) -- (1.2,-3.8) -- cycle;
\node[inner sep=0pt] at (0.7,-4.3) 
    {\includegraphics[width=.055\textwidth,angle=-90]{acquisition/slice_ax.png}};
\draw[fill=cyan] (0.1,-3.9) -- (0.1,-4.9) -- (1.1,-4.9) -- (1.1,-3.9) -- cycle;
\node[inner sep=0pt] at (0.6,-4.4) 
    {\includegraphics[width=.055\textwidth,angle=-90]{acquisition/slice_ax.png}};
\draw[fill=cyan] (0,-5) -- (0,-4) -- (1,-4) -- (1,-5) -- cycle;
\node[inner sep=0pt] at (0.5,-4.5) 
    {\includegraphics[width=.055\textwidth,angle=-90]{acquisition/slice_ax.png}};
\draw[fill=cyan] (-0.1,-5.1) -- (-0.1,-4.1) -- (0.9,-4.1) -- (0.9,-5.1) -- cycle;
\node[inner sep=0pt] at (0.4,-4.6) 
    {\includegraphics[width=.055\textwidth,angle=-90]{acquisition/slice_ax.png}};
\draw[fill=cyan] (-0.2,-5.2) -- (-0.2,-4.2) -- (0.8,-4.2) -- (0.8,-5.2) -- cycle;
\node[inner sep=0pt] at (0.3,-4.7) 
    {\includegraphics[width=.055\textwidth,angle=-90]{acquisition/slice_ax.png}};
\draw[->]  (1,-5.2) -- (1.4,-4.8);
\draw[->]  (-.2,-5.35) -- (.8,-5.35);
\draw[->]  (-0.35,-5.2) -- (-0.35,-4.2);
\node at (.3,-5.55) {$x$};
\node at (-0.55,-4.7) {$y$};
\node at (1.3,-5.1) {$t$};

\node at (1.6,-4.5) {$\dots$};

\draw[fill=magenta] (2.4,-3.8) -- (2.4,-4.8) -- (3.4,-4.8) -- (3.4,-3.8) -- cycle;
\node[inner sep=0pt] at (2.9,-4.3) 
    {\includegraphics[width=.055\textwidth,angle=-90]{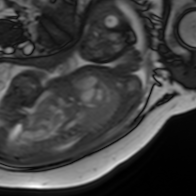}};
\draw[fill=magenta] (2.3,-3.9) -- (2.3,-4.9) -- (3.3,-4.9) -- (3.3,-3.9) -- cycle;
\node[inner sep=0pt] at (2.8,-4.4) 
    {\includegraphics[width=.055\textwidth,angle=-90]{acquisition/slice_sag.png}};
\draw[fill=magenta] (2.2,-5) -- (2.2,-4) -- (3.2,-4) -- (3.2,-5) -- cycle;
\node[inner sep=0pt] at (2.7,-4.5) 
    {\includegraphics[width=.055\textwidth,angle=-90]{acquisition/slice_sag.png}};
\draw[fill=magenta] (2.1,-5.1) -- (2.1,-4.1) -- (3.1,-4.1) -- (3.1,-5.1) -- cycle;
\node[inner sep=0pt] at (2.6,-4.6) 
    {\includegraphics[width=.055\textwidth,angle=-90]{acquisition/slice_sag.png}};
\draw[fill=magenta] (2,-5.2) -- (2,-4.2) -- (3,-4.2) -- (3,-5.2) -- cycle;
\node[inner sep=0pt] at (2.5,-4.7) 
    {\includegraphics[width=.055\textwidth,angle=-90]{acquisition/slice_sag.png}};

\node at (3.8,-4.5) {$\dots$};

\draw[fill=teal] (4.6,-3.8) -- (4.6,-4.8) -- (5.6,-4.8) -- (5.6,-3.8) -- cycle;
\node[inner sep=0pt] at (5.1,-4.3) 
    {\includegraphics[width=.055\textwidth,angle=-90]{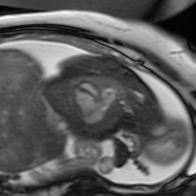}};
\draw[fill=teal] (4.5,-3.9) -- (4.5,-4.9) -- (5.5,-4.9) -- (5.5,-3.9) -- cycle;
\node[inner sep=0pt] at (5.0,-4.4) 
    {\includegraphics[width=.055\textwidth,angle=-90]{acquisition/slice_cor.png}};
\draw[fill=teal] (4.4,-5) -- (4.4,-4) -- (5.4,-4) -- (5.4,-5) -- cycle;
\node[inner sep=0pt] at (4.9,-4.5) 
    {\includegraphics[width=.055\textwidth,angle=-90]{acquisition/slice_cor.png}};
\draw[fill=teal] (4.3,-5.1) -- (4.3,-4.1) -- (5.3,-4.1) -- (5.3,-5.1) -- cycle;
\node[inner sep=0pt] at (4.8,-4.6) 
    {\includegraphics[width=.055\textwidth,angle=-90]{acquisition/slice_cor.png}};
\draw[fill=teal] (4.2,-5.2) -- (4.2,-4.2) -- (5.2,-4.2) -- (5.2,-5.2) -- cycle;
\node[inner sep=0pt] at (4.7,-4.7) 
    {\includegraphics[width=.055\textwidth,angle=-90]{acquisition/slice_cor.png}};

\node at (2.7,-6.1) {Stacks of gated 2D$+$time slices};
\node at (2.7,-6.45) {in multiple orientations};

\draw[->]  (6,-4.5) -- (8,-4.5);
\node at (7,-4.15) {\textbf{SPARC}};

\draw[fill=gray] (8.6,-5.2) -- (8.6,-4.2) -- (9.6,-4.2) -- (9.6,-5.2) -- cycle;
\node[inner sep=0pt] at (9.1,-4.7) 
    {\includegraphics[width=.055\textwidth]{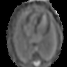}};
\draw[fill=gray] (9,-3.8) -- (8.6,-4.2) -- (9.6,-4.2) -- (10,-3.8) -- cycle;
\draw[fill=gray] (10,-4.8) -- (9.6,-5.2) -- (9.6,-4.2) -- (10,-3.8) -- cycle;
\draw[->]  (9.8,-5.2) -- (10.2,-4.8);
\draw[->]  (8.6,-5.35) -- (9.6,-5.35);
\draw[->]  (8.45,-5.2) -- (8.45,-4.2);
\node at (9.1,-5.55) {$x$};
\node at (8.25,-4.7) {$y$};
\node at (10.1,-5.1) {$z$};

\node at (10.4,-4.5) {$\dots$};

\draw[fill=gray] (10.8,-5.2) -- (10.8,-4.2) -- (11.8,-4.2) -- (11.8,-5.2) -- cycle;
\node[inner sep=0pt] at (11.3,-4.7) 
    {\includegraphics[width=.055\textwidth]{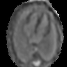}};
\draw[fill=gray] (11.2,-3.8) -- (10.8,-4.2) -- (11.8,-4.2) -- (12.2,-3.8) -- cycle;
\draw[fill=gray] (12.2,-4.8) -- (11.8,-5.2) -- (11.8,-4.2) -- (12.2,-3.8) -- cycle;

\node at (12.6,-4.5) {$\dots$};

\draw[fill=gray] (13,-5.2) -- (13,-4.2) -- (14,-4.2) -- (14,-5.2) -- cycle;
\node[inner sep=0pt] at (13.5,-4.7) 
    {\includegraphics[width=.055\textwidth]{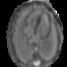}};
\draw[fill=gray] (13.4,-3.8) -- (13,-4.2) -- (14,-4.2) -- (14.4,-3.8) -- cycle;
\draw[fill=gray] (14.4,-4.8) -- (14,-5.2) -- (14,-4.2) -- (14.4,-3.8) -- cycle;

\draw[->]  (8.6,-6) -- (14.4,-6);
\draw  (9.1,-6.2) -- (9.1,-5.8);
\draw  (11.3,-6.2) -- (11.3,-5.8);
\draw  (13.5,-6.2) -- (13.5,-5.8);
\node at (14.4,-5.7) {$t$};
\node at (11.25,-6.5) {Reoriented cine with $T$ cardiac phases};

\end{tikzpicture}
\caption{Multiple gated 2D$+$time stacks acquired in orthogonal orientations are combined via the \textbf{SPARC} pipeline to produce a 3D$+$time cine volume with $T$ cardiac phases reoriented in canonical fetal coordinate system.}
\label{fig:pipeline_intro}
\end{figure*}

In this work, we propose \textbf{S}lice-to-volume \textbf{P}ipeline for \textbf{A}utomated \textbf{R}econstruction of gated 3D$+$time fetal \textbf{C}ardiac MRI (\textbf{SPARC}), the first automated end-to-end pipeline for localisation, reconstruction and reorientation of 3D$+$time fetal cardiac cine volumes from Doppler-gated MRI stacks (Figure~\ref{fig:pipeline_intro}). This hybrid pipeline combines a physics-informed slice-to-volume reconstruction algorithm with deep learning models for thoracic segmentation and anatomical reorientation. The main contributions of this work are as follows:
\begin{enumerate}
    \item A parsimonious slice-wise spatio-temporal forward model specifically adapted to Doppler-gated fetal cardiac cine reconstruction achieving a tenfold reduction in reconstruction time relative to frame-wise approaches \citep{van_amerom_fetal_2019} with improved generalisation to unseen data.

    \item A transfer learning strategy for  training deep learning models (i.e., thoracic segmentation, cardiac centre localisation, and reorientation) for Doppler-gating data, to take advantage of a larger free-running domain.

    \item An end-to-end clinically validated pipeline, packaged as a Docker container and currently deployed at our institution, evaluated on a large held-out clinical cohort ($n = 121$) with $82.6\%$ fully automatic success and a mean processing time of $7.1 \pm 1.3$ min compatible with clinical deployment requirements.
\end{enumerate}

\section{Methodology}

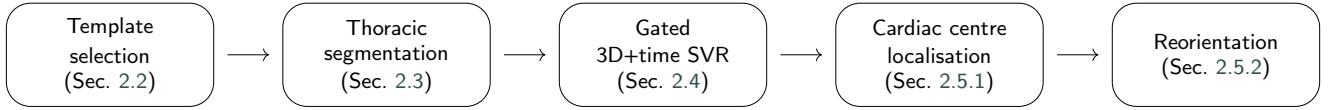
\begin{figure*}[t]
\centering
\begin{adjustbox}{width=\textwidth}
\begin{tikzpicture}[every node/.style={inner sep=0,outer sep=0}]

\draw[rounded corners=10] (0,0) -- (3,0) -- (3,1.5) -- (0,1.5) -- cycle;
\node at (1.5,1.15) {Template};
\node at (1.5,0.75) {selection};
\node at (1.5,0.35) {(Sec.~\ref{sec:template})};

\draw[->] (3.2,0.75) -- (3.8,0.75);
\draw[rounded corners=10] (4,0) -- (7,0) -- (7,1.5) -- (4,1.5) -- cycle;
\node at (5.5,1.15) {Thoracic};
\node at (5.5,0.75) {segmentation};
\node at (5.5,0.35) {(Sec.~\ref{sec:segmentation})};

\draw[->] (7.2,0.75) -- (7.8,0.75);
\draw[rounded corners=10] (8,0) -- (11,0) -- (11,1.5) -- (8,1.5) -- cycle;
\node at (9.5,1.15) {Gated};
\node at (9.5,0.75) {3D$+$time SVR};
\node at (9.5,0.35) {(Sec.~\ref{sec:recon})};

\draw[->] (11.2,0.75) -- (11.8,0.75);
\draw[rounded corners=10] (12,0) -- (15,0) -- (15,1.5) -- (12,1.5) -- cycle;
\node at (13.5,1.15) {Cardiac centre};
\node at (13.5,0.75) {localisation};
\node at (13.5,0.35) {(Sec.~\ref{sec:heart_segmentation})};

\draw[->] (15.2,0.75) -- (15.8,0.75);
\draw[rounded corners=10] (16,0) -- (19,0) -- (19,1.5) -- (16,1.5) -- cycle;
\node at (17.5,0.95) {Reorientation};
\node at (17.5,0.55) {(Sec.~\ref{sec:transformer-based_reorientation})};

\end{tikzpicture}
\end{adjustbox}
\caption{Overview of the proposed \textbf{SPARC} pipeline. }
\label{fig:pipeline}
\end{figure*}

The proposed \textbf{SPARC} pipeline consists of three fully automated steps that transform stacks of thick, gated 2D$+$time slices into an isotropic cine volume of the fetal heart, reoriented into a standard fetal coordinate space (Figure~\ref{fig:pipeline}). The steps are: (1) template stack selection and segmentation of fetal thorax (Sec.~\ref{sec:template} $\&$ Sec.~\ref{sec:segmentation}), (2) reconstruction of fetal heart 3D$+$time volume spanning a single cardiac cycle (Sec.~\ref{sec:recon}), and (3) reorientation to the fetal anatomical coordinate system (Sec.~\ref{sec:reo}). Before describing these modules, we briefly review Doppler-ultrasound gated MR acquisition and introduce the mathematical notation used in this work.

\subsection{Doppler-gated MR acquisition}

During acquisition, the Doppler-ultrasound device measures fetal cardiac wall motion and flow velocity to derive a gating signal transmitted to the MRI scanner \citep{kording_evaluation_2018, kording_dynamic_2018}. During the acquisition of each 2D$+$time slice, the Doppler device triggers acquisition of part of the $k$-space at each cardiac cycle (Figure~\ref{fig:doppler_acquisition}).
Since the cardiac cycle duration varies across heartbeats due to physiological heart rate variability, the number of acquired cardiac phases per cycle is not constant. Temporal interpolation in $k$-space to a fixed number of reconstructed frames $T$ is therefore applied. Ultimately, each slice spans a single cardiac cycle, with each frame composed from $k$-space samples collected over multiple heartbeats, and there is a one-to-one correspondence between temporal indices $t \in \{1, \dots, T \}$ and cardiac phases.

In this study, multiple stacks $\mathbf{s}_l$ of thick, Doppler-gated 2D$+$time slices $\mathbf{y}_{k}$ covering the fetal thorax were acquired in different orientations (i.e., axial, sagittal, coronal, and oblique). Let $\mathbf{y}_{k} = (\mathbf{y}_{k,1}^{\top}, \dots, \mathbf{y}_{k,T}^{\top})^{\top} \in \mathbb{R}^{N_k\times T}$ denote the vector representation of the $k$-th slice, composed of $T$ frames with $N_k$ voxels each, and $\mathbf{s}_l = \{ \mathbf{y}_k \mid k \in \mathcal{E}_l \}$ denote the $l$-th stack, where $\mathcal{E}_l \subset \{1, \dots, K\}$ is the index set of slices belonging to that stack. 

As a preprocessing step, all stacks were corrected for slice wise bias field inhomogeneity using the N4 algorithm \citep{tustison_n4itk_2010}. Owing to the gated acquisition, the bias field was estimated from the first frame and the resulting correction was applied to all frames.

\subsection{Motion-robust template selection}
\label{sec:template}

The axial stack exhibiting the lowest level of motion artefacts was automatically selected as the reconstruction template. Motion in each time-averaged stack $\mathbf{\bar{s}}_l$ was evaluated as the normalised variance of adjacent slice differences: 
\begin{equation}
    \Delta_l^2 = \dfrac{\textrm{Var}(\{ \| \mathbf{\bar{y}}_k - \mathbf{\bar{y}}_{k+1}\|_1 \mid (k, k+1) \in \mathcal{E}_l^2 \})}{(\textrm{Avg}(\{ \|\mathbf{\bar{y}}_k\|_1 \mid k \in \mathcal{E}_l\}))^2}
\end{equation}
This variance-based $z$-smoothness metric captures the instability between neighbouring slices induced by fetal motion. A low $\Delta_l^2$ indicates stable fetal positioning throughout the stack, which is the primary requirement for a reliable reconstruction template.

\subsection{Thoracic segmentation via transfer learning}
\label{sec:segmentation}

Automatic thoracic segmentation was subsequently performed on the chosen template stack to define the region of interest for reconstruction. A lightweight 3D UNet \citep{ronneberger_u-net_2015} was trained on stacks acquired in axial, coronal, sagittal and oblique directions. Training incorporated extensive data augmentation, including random translations, rotations (in-plane and through-plane), scaling, and flipping, as well as intensity-based gamma transformation and bias-field augmentation. The agreement between predicted and manually annotated ground-truth masks was measured using an equally weighted combination of Dice and cross-entropy (CE) losses \citep{cardoso_monai_2022}.

Due to the scarcity of annotated Doppler-gated data, a transfer learning strategy was adopted. The network was first trained on a large-scale free-running (i.e., non-gated) fetal cardiac dataset acquired on a different scanner (source domain $\mathcal{D}_S$), and subsequently fine-tuned on the target domain $\mathcal{D}_T$ (i.e., Doppler-gated) using transfer learning \citep{guan_domain_2022}. 

An ensemble strategy was adopted to enhance robustness and generalizability \citep{cardoso_monai_2022}. Networks were trained on distinct splits of the training data, and their predictions were aggregated at inference via majority voting. The largest connected component was retained, and residual mask holes were filled to promote anatomical plausibility \citep{cardoso_monai_2022}. A final binary morphological dilation ($1\times$voxel size in-plane) was applied to the thorax mask to provide a safety margin against boundary uncertainty, reducing the risk of excluding cardiac structures during reconstruction when the heart lies adjacent to the thoracic wall.
    
\subsection{Motion-corrected cine reconstruction}
\label{sec:recon}
\begin{figure*}[t]
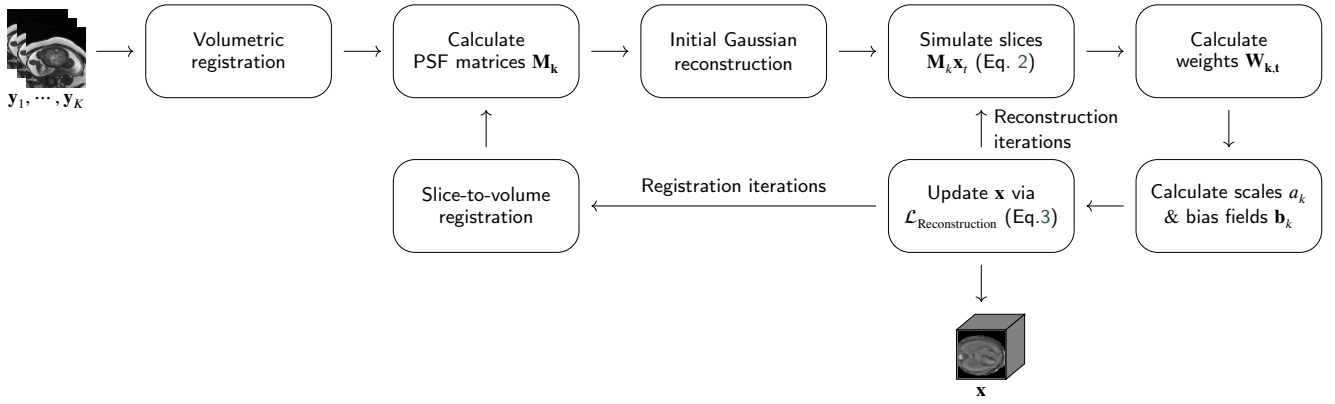

\centering
\begin{adjustbox}{width=\textwidth}
\begin{tikzpicture}[every node/.style={inner sep=0,outer sep=0}]

\node[inner sep=0pt] at (-1.75,0.95) 
    {\includegraphics[width=0.055\textwidth, angle=-90]{acquisition/slice_ax.png}};
\node[inner sep=0pt] at (-1.6,0.80)
    {\includegraphics[width=0.055\textwidth, angle=-90]{acquisition/slice_ax.png}};
\node[inner sep=0pt] at (-1.45,0.65)
    {\includegraphics[width=0.055\textwidth, angle=-90]{acquisition/slice_ax.png}};
\node at (-1.6,-0.05) {$\mathbf{y}_{1}, \cdots, \mathbf{y}_{K}$};
\draw[->] (-0.8,0.75) -- (-0.2,0.75);

\draw[rounded corners=10] (0,0) -- (3,0) -- (3,1.5) -- (0,1.5) -- cycle;
\node at (1.5,.95) {Volumetric};
\node at (1.5,0.55) {registration};

\draw[->] (3.2,0.75) -- (3.8,0.75);
\draw[rounded corners=10] (4,0) -- (7,0) -- (7,1.5) -- (4,1.5) -- cycle;
\node at (5.5,0.95) {Calculate};
\node at (5.5,0.55) {PSF matrices $\mathbf{M_k}$};

\draw[->] (7.2,0.75) -- (7.8,0.75);
\draw[rounded corners=10] (8,0) -- (11,0) -- (11,1.5) -- (8,1.5) -- cycle;
\node at (9.5,0.95) {Initial Gaussian};
\node at (9.5,0.55) {reconstruction};

\draw[->] (11.2,0.75) -- (11.8,0.75);
\draw[rounded corners=10] (12,0) -- (15,0) -- (15,1.5) -- (12,1.5) -- cycle;
\node at (13.5,0.95) {Simulate slices};
\node at (13.5,0.55) {$\mathbf{M}_{k}\mathbf{x}_{t}$ (Eq. \ref{eq:acquisitionmodel})};

\draw[->] (15.2,0.75) -- (15.8,0.75);
\draw[rounded corners=10] (16,0) -- (19,0) -- (19,1.5) -- (16,1.5) -- cycle;
\node at (17.5,0.95) {Calculate};
\node at (17.5,0.55) {weights $\mathbf{W_{k,t}}$};

\draw[->] (17.5,-.2) -- (17.5,-.8);
\draw[rounded corners=10] (16,-1) -- (19,-1) -- (19,-2.5) -- (16,-2.5) -- cycle;
\node at (17.5,-1.55) {Calculate scales $a_k$ };
\node at (17.5,-1.95) {$\&$ bias fields $\mathbf{b}_k$};

\draw[->] (15.8,-1.75) -- (15.2,-1.75);
\draw[rounded corners=10] (12,-1) -- (15,-1) -- (15,-2.5) -- (12,-2.5) -- cycle;
\node at (13.5,-1.55) {Update $\mathbf{x}$ via};
\node at (13.5,-1.95) {$\mathcal{L}_{\textrm{Reconstruction}}$ (Eq.\ref{eq:recon_loss})};

\draw[->] (13.5,-.8) -- (13.5,-.2);
\node[anchor=west] at (13.7,-.3) {Reconstruction};
\node[anchor=west] at (13.7,-.7) {iterations};

\draw[->] (11.8,-1.75) -- (7.2,-1.75);
\node at (9.5,-1.45) {Registration iterations};

\draw[->] (5.5,-.8) -- (5.5,-.2);
\draw[rounded corners=10] (4,-1) -- (7,-1) -- (7,-2.5) -- (4,-2.5) -- cycle;
\node at (5.5,-1.55) {Slice-to-volume};
\node at (5.5,-1.95) {registration};

\draw[fill=gray] (13.1,-4.55) -- (13.1,-3.75) -- (13.9,-3.75) -- (13.9,-4.55) -- cycle;
\node[inner sep=0pt] at (13.5,-4.15)
    {\includegraphics[width=0.0425\textwidth, angle=-90]{acquisition/vol_t0_ax.png}};
\draw[fill=gray] (13.1,-3.75) -- (13.4,-3.5) -- (14.2,-3.5) -- (13.9,-3.75) -- cycle;
\draw[fill=gray] (13.9,-4.55) -- (14.2,-4.3) -- (14.2,-3.5) -- (13.9,-3.75) -- cycle;
\node at (13.5,-4.75) {$\mathbf{x}$};
\draw[->] (13.5,-2.7) -- (13.5,-3.3);

\end{tikzpicture}
\end{adjustbox}
\caption{Overview of the slice-to-volume reconstruction algorithm.}
\label{fig:svr}
\end{figure*}

\subsubsection{Spatio-temporal forward model}
\label{sec:forward}

Acquired stacks of 2D$+$time slices are affected by inter-slice motion, as the fetus is moving freely and the mother is breathing normally. Fetal motion is assumed to be rigid since the fetal thorax is contained within the rib-cage, and the whole 2D$+$time slice is assumed to be in the same motion-state, because Fourier samples are acquired continuously and then retrospectively binned into cardiac phases.

The forward model is a linear matrix operation that simulates intensity-corrected  frames $\mathbf{y}_{k,t}^*$ from the 3D$+$time reconstructed cardiac anatomy 
$\mathbf{x}\in \mathbb{R}^{N\times T}$  
\begin{equation}
    \mathbf{y}_{k,t}^*=\mathbf{M}_{k}\mathbf{x}_{t};\quad \mathbf{y}_{k,t}^*=a_{k}\exp(-\mathbf{b}_k)\circ\mathbf{y}_{k,t}
    \label{eq:acquisitionmodel}
\end{equation}
where $\mathbf{x}_{t}$ is a reconstructed 3D frame corresponding to cardiac phase $t$. The sparse matrix $\mathbf{M}_{k}\in \mathbb{R}^{N_k\times N}$ is composed of rows of vectorised 3D point spread functions (PSF) corresponding to voxels of the acquired slice $\mathbf{y}_{k}$, reoriented using motion parameters and sampled on the grid of reconstructed image $\mathbf{x}$. Note that PSF is not dependent on time $t$ and is applied to all frames equally. The PSF is determined by the acquisition protocol and it is approximated by a truncated 3D Gaussian function with full width at half maximum equal to the slice-thickness in the through-plane direction and  $1.2\times$voxel size in-plane respectively \citep{kuklisova-murgasova_reconstruction_2012}. 

Intensity correction is modeled by applying of slice-wise scales $a_k$ and spatially smooth slice-wise bias fields $\mathbf{b}_k$, where $\exp(.)$ is applied element wise to the vector $-\mathbf{b}_k$ and $\circ$ denotes element wise multiplication. These intensity correction parameters model interactions of MRI signal with fetal motion, such as partial slice dropouts, spin history effects and differential transmit and receive MRI field inhomogeneities \citep{kuklisova-murgasova_reconstruction_2012}. These effects are assumed to be independent of the cardiac phase $t$ and are applied to all frames of the 2D$+$time slices equally.

\subsubsection{Slice-to-volume motion estimation}
\label{sec:MC}

Motion correction is initialised by volumetric registration to align each stack $\mathbf{s}_l$ to the previously selected template stack (Sec.~\ref{sec:template}), enabling an initial volume reconstruction, followed by iterations of slice-to-volume registration interleaved with super-resolution reconstruction (Fig.~\ref{fig:svr}). Rigid transformations with normalized mutual information similarity metric were adopted for both stages \citep{kuklisova-murgasova_reconstruction_2012, van_amerom_fetal_2018}. Given the gated acquisition, all frames within a slice are assumed to share a common motion state. Consequently, a single 3D rigid transformation estimated from a single temporal frame is sufficient to align the entire set of frames associated with that slice.

\subsubsection{Spatio-temporal super-resolution reconstruction}
\label{sec:SR}

Assuming that the spatial alignment, and consequently also PSF matrices $\mathbf{M}_k$ are known, the cine volume $\mathbf{x}$ can be estimated by minimizing the weighted sum of squared residuals between the intensity-corrected \textit{acquired} slices $\mathbf{y}_{kt}^*$ and \textit{simulated} slices $\mathbf{M}_{k}\mathbf{x}_{t}$ (Eq.~\ref{eq:acquisitionmodel}) to ensure their consistency assumed in the forward model:
\begin{equation}
\label{eq:recon_loss}
    \mathcal{L}_{\textrm{Reconstruction}}(\mathbf{x}) =  \sum_{k=1}^{K} \sum_{t=1}^{T} \|\mathbf{W}_{k,t} (\mathbf{y}_{k,t}^* - \mathbf{M}_k \mathbf{x}_t) \|_2^2 + \lambda \mathcal{R}(\mathbf{x})
\end{equation}
where $\mathbf{W}_{kt}$ are diagonal matrices with slice-voxel level weights $w_{j,k,t} = q_k p_{j,k,t}$ with $p_{j,k,t}$ representing voxel-level and $q_k$ slice-level confidence in the acquired data.  

Robust statistics matrices $\mathbf{W}_{k,t}$ were estimated within an expectation–maximization (EM) framework following \citep{kuklisova-murgasova_reconstruction_2012, van_amerom_fetal_2018}, by classification of each voxel and slice as an inlier or outlier based on the error terms $\mathbf{y}_{k,t}^* - \mathbf{M}_k \mathbf{x}_t$, thus rejecting corrupted and misaligned slices from the reconstruction. 

Edge-preserving regularization $\mathcal{R}(.)$ \citep{charbonnier_deterministic_1997, kuklisova-murgasova_reconstruction_2012} was applied independently to each 3D cardiac frame to stabilize the reconstruction while limiting over-smoothing of anatomical structures. The regularization term was computed for each voxel based on its 26 spatial neighbours. Parameter $\lambda$ controls the strength of regularization, and was set to $\lambda = 0.02$.

The intensity correction parameters $\mathbf{b}_k$ and $a_k$ are also estimated by minimising the loss (Eq.~\ref{eq:recon_loss}) jointly with the reconstruction in an interleaved manner (Fig.~\ref{fig:svr}) following the approach of \cite{kuklisova-murgasova_reconstruction_2012}.

As a post-processing step, the cine volume is corrected for volumetric bias field inhomogeneity using the N4 algorithm \citep{tustison_n4itk_2010}. Given the temporal consistency afforded by gated acquisition, the bias field was estimated from the first frame and propagated to all frames.

\subsection{Anatomical reorientation}
\label{sec:reo}

The position and orientation of the fetus in the template stack, and consequently in the reconstructed volume $\mathbf{x}$, are unpredictable \textit{a priori}. The cine volume must therefore be reoriented to a canonical fetal coordinate system to enable consistent anatomical interpretation and quantitative analysis. The proposed anatomical reorientation process was decomposed into two steps: (1) alignment to the fetal heart centre of mass to correct large translations (Sec.~\ref{sec:heart_segmentation}), and (2) reorientation to the fetal coordinate system using a Transformer network (sec.~\ref{sec:transformer-based_reorientation}). The two stages were performed on the time-averaged reconstructed volume $\mathbf{\bar{x}}$ to improve signal-to-noise ratio (SNR) and enhance the robustness of anatomical landmark detection. $\mathbf{\bar{x}}$ is treated in this section as a 3D image (not a flattened vector).

\subsubsection{Cardiac centre localization}
\label{sec:heart_segmentation}
We employed the same transfer-learning strategy to train a 3D U-Net segmentation of the fetal heart as in Sec.~\ref{sec:segmentation}. Extensive intensity and spatial augmentations, including full 3D rotations, were applied during training to promote generalisability. During inference, predictions from separate heart-segmentation networks trained on distinct data splits were combined using a majority voting strategy. Anatomical plausibility of the predicted heart mask was promoted via largest connected component selection, and residual mask holes filling. Fetal heart centre of mass was extracted from the predicted mask and used to recentre the image prior to reorientation. 

\subsubsection{Transformer-based 3D reorientation}
\label{sec:transformer-based_reorientation}

\paragraph{Anchor point representation.} Reorientation was formulated as an image regression task, where predicted transformations were parametrised using 
anchor-points similarly to \citet{hou_3-d_2018}. Three anchor points were defined: the image centre (corresponding to the fetal cardiac centre), and the left- and right-posterior midpoints of the image. These non-collinear points uniquely determine a rigid transformation. An orthonormal reference frame was constructed via Gram–Schmidt orthogonalization from the predicted anchors, from which the translation vector and rotation matrix were recovered. The obtained transformation enables anatomical reorientation of the image to the canonical fetal coordinate system.

\paragraph{Training strategy.} We employed a standard vision Transformer (ViT) architecture to predict the spatial orientation. Manually curated fetal heart volumes in a canonical coordinate system were used for training. To simulate arbitrary rigid poses, random translations and rotations were applied, and the network was trained to predict the corresponding anchor-point locations in the transformed volume \citep{hou_3-d_2018, hou_computing_2018}. Translations were sampled uniformly within $20\%$ of the image size along each axis. This restricted range was sufficient due to previous centering step. Rotations were sampled over $\mathrm{SO}(3)$ using an Euler-angle parameterization, with $\theta \sim \mathcal{U}(0, 2\pi)$, $\cos(\varphi) \sim \mathcal{U}(-1,1)$ and $\psi \sim \mathcal{U}(-\pi, \pi)$, ensuring uniform coverage of 3D orientations. 

To promote invariance to mask shape and background context, mask-based augmentations were applied, including random erosion of the thorax mask and random dilation of the heart mask. Intensity augmentations, such as contrast perturbation and simulated bias-field inhomogeneity, were also employed to improve robustness and generalization. 

\paragraph{Loss function and optimisation.} The network parameters 
$\mathbf{\Theta}_{\textrm{ViT}}$ were optimized using a weighted combination of four complementary loss terms:
\begin{align}
\begin{split}
    \mathcal{L}_{\textrm{Reorientation}}(\mathbf{\Theta}_{\textrm{ViT}}) &= \lambda_1 \mathcal{L}_{\textrm{Anchor}} + \lambda_2 \mathcal{L}_{\textrm{Geodesic}} \\
    &+ \lambda_3 \mathcal{L}_{\textrm{Translation}} + \lambda_4 \mathcal{L}_{\textrm{Image}}
\end{split}
\end{align}
Here, $\mathcal{L}_{\textrm{Anchor}}$ denotes the Euclidean distance between predicted and ground-truth anchor points; $\mathcal{L}_{\textrm{Geodesic}}$ is the geodesic distance between the predicted and ground-truth rotation matrices; $\mathcal{L}_{\textrm{Translation}}$ measures the Euclidean distance between predicted and ground-truth translations; and $\mathcal{L}_{\textrm{Image}}$ corresponds to the $L^1$ norm between the reoriented and canonical images. The geodesic distance was defined as the angle $\arccos\!\left(\frac{\mathrm{tr}(\mathbf{R}^\top \hat{\mathbf{R}}) - 1}{2}\right)$ between the predicted and ground-truth rotation matrices $\hat{\mathbf{R}}$ and $\mathbf{R}$ \citep{hartley_rotation_2013}. The proposed multi-objective loss enforces consistency at both the parameter and image levels. Furthermore, $\mathcal{L}_{\textrm{Anchor}}$ provides spatial supervision of the anchor configuration, while explicit geodesic and translation losses regularize the induced rigid transformation in $\mathrm{SE}(3)$, stabilizing training and preventing degenerate anchor configurations. Optimization was carried out via the Adam algorithm \citep{kingma_adam_2017} and the hyperparameters were set empirically to: $\lambda_1 = 0.01$, $\lambda_2 = 0.1$, $\lambda_3 = 0.01$, and $\lambda_4 = 0.1$. 

\paragraph{Ensembling.} Separate networks, initialized via transfer learning from $\mathcal{D}_S$, were trained on distinct $\mathcal{D}_T$ data splits, and an ensemble strategy was adopted at inference. For each model, the predicted anchor points were converted into the corresponding translation vector and rotation matrix. The translation vectors were aggregated by simple arithmetic averaging, while the rotation matrices were combined using chordal rotation averaging \citep{hartley_rotation_2013}, defined as:
\begin{equation}
    \mathbf{\bar{R}}_{\textrm{ViT}} = \operatorname*{arg\,min}_{\mathbf{R}\in \mathrm{SO}(3)} \sum_i \| \mathbf{R} - \mathbf{R}_{\textrm{ViT}_i} \|^2_F
\label{eq_chordal}
\end{equation}
where $\|. \|_F$ is the Frobenius norm. 

This ensemble aggregation reduced variance in reorientation while preserving geometric consistency in $\mathrm{SO}(3)$. The optimisation problem in Eq.~\ref{eq_chordal} is equivalent to finding the rotation matrix nearest to the arithmetic mean of the individual rotation matrices, and admits a closed-form solution analogous to the orthogonal Procrustes problem via singular value decomposition (SVD) \citep{hartley_rotation_2013}. Chordal rotation averaging assumes that the predictions $\mathbf{R}_{\textrm{ViT}_i}$ are relatively concentrated, which holds when the input is unambiguous. In contrast, predictions tend to be more dispersed for scans with unusual fetal anatomy or intensity distribution. Although various rotation-averaging strategies exist in the literature \citep{hartley_rotation_2013}, we adopted chordal averaging as our ensemble aggregation strategy, as empirically validated in Section~\ref{sec:results_reorientation}.

\section{Experiments}

\subsection{Fetal cardiac MRI datasets}

\subsubsection{Doppler-gated dataset \--- Target domain $\mathcal{D}_T$}

Cine stacks of 2D$+$time slices were acquired from $199$ subjects (gestational age: $32.5 \pm 1.5$ [$27.4$, $37.0$] weeks), comprising $18$ healthy volunteers and $181$ fetuses with congenital heart disease and related conditions, on a 1.5T MAGNETOM Sola (Siemens Healthineers, Germany) using Doppler ultrasound gated \citep{kording_evaluation_2018, kording_dynamic_2018} balanced steady-state free precession (bSSFP) sequence (TE $1.93$ ms, TR $56.3$ ms, flip angle $60\degree$, and acceleration factor $3$, spatial resolution $2 \times 2 \times 6$ mm$^3$, $T = 20$ temporal frames after resampling), see Figure~\ref{fig:doppler_acquisition}. Four to 16 stacks were acquired in multiple orientations. Each 2D$+$time slice spanned a single cardiac cycle, with a per-slice acquisition time of approximately $4$ seconds.

Manual thoracic and cardiac segmentation masks, along with canonical reorientation, were produced using ITK-SNAP \citep{yushkevich_user-guided_2006}.The first rater annotated both the training and test sets. A second rater independently annotated the test data in order to assess inter-rater agreement. The \textbf{SPARC} pipeline was additionally evaluated on a large held-out subset not used at any stage of network training or evaluation, for which no ground truth annotations were available. Dataset splits are described in Table~\ref{table:data-split}.  

\subsubsection{Free-running dataset \--- Source domain $\mathcal{D}_S$}

Free-running cine stacks of 2D$+$time slices (4-8 stacks in multiple orientations) were acquired from $141$ subjects (gestational age: $32.3 \pm 1.7$ [$27.9$, $39.3$] weeks), comprising $6$ healthy volunteers and $135$ fetuses with congenital heart disease and related conditions, on a 1.5T Ingenia (Philips Healthcare, Netherlands) using a 2D bSSFP sequence (TE $1.9$ ms, TR $3.8$ ms, flip angle $60\degree$, spatial resolution $2 \times 2 \times 6$ mm$^3$, slice overlap 2‐3 mm, under-sampling factor of 8, temporal resolution 72 ms, 96 frames per slice) and reconstructed using a $k$-$t$ SENSE which included motion correction, cardiac synchronization using image-base gating and volumetric reconstruction \citep{van_amerom_fetal_2018, van_amerom_fetal_2019, roberts_fetal_2020}. Acquisition of a typical stack took 155s. Manual cardiac and thoracic segmentation masks, were available for all subjects. All cine volumes were manually reoriented to a canonical coordinate system using ITK-SNAP \citep{yushkevich_user-guided_2006}.

\subsection{Experimental setup}

\begin{table}[t!]
\centering
\resizebox{\columnwidth}{!}{\begin{tabular}{| c | c | c | c | c |}
\hline
\multirow{3}{*}{Domain} & \multirow{3}{*}{Evaluation} & \multicolumn{3}{|c|}{Subjects $n$} \\\cline{3-5}
& & \multirow{2}{*}{\shortstack{5-fold cross-\\validation set}} & \multirow{2}{*}{\shortstack{Held-out\\test set}} & \multirow{2}{*}{\shortstack{Unique\\subjects}}\\
& & & & \\ \hline\hline
\multirow{3}{*}{Source} & Thoracic segmentation & 116 [483] & 23 (92) & \multirow{3}{*}{141} \\ \cline{2-4}
& Cardiac segmentation & 98 & 22 & \\ \cline{2-4}
& Reorientation & 91 & 21 & \\ \hline\hline

\multirow{6}{*}{Target} & Thoracic segmentation & 25 [222] & 5 (36) & \multirow{4}{*}{60}\\ \cline{2-4}
& Reconstruction & \--- & 10 & \\ \cline{2-4}
& Cardiac segmentation & 25 & 5 & \\ \cline{2-4}
& Reorientation & 32 & 7 & \\ \cline{2-5}
& End-to-end pipeline & \--- & 121 & 121 \\ \cline{2-5}
& Excluded & \--- & \--- & 18 \\ \hline
\end{tabular}}

\caption{Dataset splits for each evaluated task in the source (free-running; $n = 141$ subjects) and target (Doppler-gated; $n = 199$ subjects) domains. Numbers of stacks are shown in brackets.}
\label{table:data-split}
\end{table}

\subsubsection{Deep learning models}

For each deep learning task (Sec.\ref{sec:segmentation}, \ref{sec:heart_segmentation} and \ref{sec:transformer-based_reorientation}), five networks were trained and validated on distinct subject-level splits of the cross-validation set (Table~\ref{table:data-split}), used for both hyperparameter tuning and model selection. The resulting five trained networks were subsequently evaluated on a held-out test set using an ensemble strategy: majority voting for segmentation tasks and chordal rotation averaging for reorientation (sec.~\ref{sec:transformer-based_reorientation}). Cross-validation and test subjects were selected to reflect the overall distribution of the cohort in terms of gestational age and cardiac condition (volunteer vs. CHD). 

We investigated four domain adaptation strategies: \textbf{(1) Source-only} networks trained exclusively on $\mathcal{D}_S$; \textbf{(2) Target-only} networks trained exclusively on $\mathcal{D}_T$; \textbf{(3) Joint} networks trained on the combined domain $\mathcal{D}_S \cup \mathcal{D}_T$; \textbf{(4) Transfer} networks pre-trained on $\mathcal{D}_S$ and subsequently fine-tuned on $\mathcal{D}_T$. All approaches were evaluated on the target test set. Approaches (1) and (3) were evaluated on the source test set, providing a benchmark in-domain performance. 

To assessed the benefit of ensemble aggregation  ensemble performance was compared against the mean performance of the five individual networks. For reorientation specifically, three ensemble aggregation strategies were compared: chordal rotation averaging, quaternion averaging, and geodesic averaging \citep{hartley_rotation_2013}.

\subsubsection{Reconstruction algorithm}
\label{sec:experiment_recon}

The proposed reconstruction algorithm (Sec.\ref{sec:recon}) was compared against the method of \citep{van_amerom_fetal_2018}, which assumes distinct motion states for each temporal frame within a slice and incorporates a $\operatorname{sinc}$ temporal point spread function (tPSF), a design choice reflecting its original development for free-running acquisitions. In contrast to the proposed approach, the method of \citep{van_amerom_fetal_2018} does not share parameters across temporal frames, resulting in a less parsimonious model with a substantially larger parameter space. The same hyperparameters were used to define the iterative reconstruction process of both methods.

The comparison was performed on $n = 10$ target domain cases  (Table~\ref{table:data-split}) using a leave-stack-out design, with a cine volume reconstructed from all but one randomly selected withheld stack. The withheld stack was subsequently simulated from the reconstructed volume and compared against the original acquired stack, assessing how well the reconstruction generalises to unseen data. Additionally, data consistency was calculated by comparing the included acquired stacks against their simulated counterparts. In both cases, comparison between acquired and simulated stacks incorporated slice registration and intensity correction to ensure a fair and consistent evaluation. To limit computational cost, a single leave-stack-out experiment was performed per subject rather than evaluating all possible stack combinations. The same stack was left-out in both methods.

\subsubsection{End-to-end pipeline}

The complete \textbf{SPARC} pipeline was evaluated on a large target domain held-out subset ($n = 139$) not used in any prior training or evaluation step. Eighteen cases were excluded on the basis of insufficient input data, falling into two categories: fewer than five stacks with adequate cardiac coverage and acceptable motion levels within stacks ($n = 14$), and loss of gating signal and consequently synchronicity between stacks ($n = 4$). These conditions prevent reliable slice-to-volume reconstruction, as sufficient spatial coverage and consistent cardiac phase synchronisation are required to recover a coherent cine volume. In $n = 6$ cases, one or more input stacks were manually excluded by the experienced observer due to excessive within-stack motion corruption. The total $n = 121$ cases were processed through the full pipeline and form the basis of the end-to-end evaluation. 

To assess the contribution of the cardiac centre localisation step, an additional ablation was performed by disabling the heart segmentation network prior to Transformer-based reorientation, and running the \textbf{SPARC} pipeline on the same $n = 121$ cases. This allowed direct comparison of reorientation performance and failure rates with and without explicit cardiac centring, as reported in Section~\ref{sec:results_pipeline}.

\subsection{Quantitative evaluation}

\subsubsection{Assessment of segmentation networks}

Segmentation performance was assessed by comparing predicted delineations against manually annotated ground-truth masks, using the Dice similarity coefficient (DSC), the $95^{\text{th}}$ percentile symmetric Hausdorff distance ($\mathrm{HD}_{95\%}$), and the average symmetric surface distance (ASSD) \citep{cardoso_monai_2022} averaged over the test set. For cardiac segmentation, the centre distance (CD) between the predicted and ground-truth centres of mass was additionally reported.

For thoracic segmentation, performance was compared across methods within each domain using a linear mixed-effects model with method as a fixed effect, subject as a random effect, and Bonferroni correction applied for multiple comparisons ($36$ stacks nested within $5$ subjects), with the proposed Transfer approach (ensemble) compared to all other methods within the target domain. Inter-rater agreement was assessed by comparing two independent sets of manual annotations on the target domain test stacks. For cardiac segmentation, given the limited size of the target domain test set ($n = 5$ subjects), statistical comparison between methods was not performed. 

\subsubsection{Assessment of reconstruction algorithm}
\label{sec:assessment_recon}

Reconstruction performance was evaluated by computing similarity metrics between the acquired and simulated stacks, measured within  manually delineated cardiac masks. Two complementary metrics were computed: normalised cross-correlation (NCC), which measures intensity agreement, and normalised root mean squared error (NRMSE), which penalises large intensity discrepancies \citep{kuklisova-murgasova_reconstruction_2012}. Both metrics were computed separately for the leave-stack-out and data consistency evaluations described in Section~\ref{sec:experiment_recon}. Reconstruction time and the proportions of included and excluded slices were recorded to characterise the computational efficiency and robust weighting behaviour of each method. Statistical comparison between methods was performed using paired $t$-tests.

\subsubsection{Assessment of reorientation network}
\label{sec:assess_reorientation}

Reorientation performance was evaluated by by applying random translations and rotations to manually reoriented fetal heart volumes (see Section~\ref{sec:transformer-based_reorientation}), enabling evaluation across the full range of possible input orientations. For each test case, $250$ random rigid transformations were sampled and held fixed across all compared methods. The error between the ground-truth and predicted rigid transformations was measured using the geodesic distance (GD) for rotations and the cardiac centre distance (CD) for translations \citep{hou_3-d_2018}. 

The proposed Transfer approach with chordal ensemble was compared to all other methods within the target domain using a linear mixed-effects model with method as a fixed effect, subject as a random effect, and Bonferroni correction applied for multiple comparisons ($1750$ samples  corresponding to 250 simulated rigid transformations per subject). Inter-rater agreement was evaluated by computing the GD between two manual transformations to a canonical orientation on the target domain test volumes.

\subsection{Qualitative evaluation of end-to-end pipeline}

\label{sec:qual_fail}
We visually evaluated each stage of the pipeline using a binary success/failure decision by two independent raters. In cases of disagreement, a decision was reached by consensus. A failure was recorded as follows: \textbf{(1) Template selection:} an axial stack exhibited less severe motion than the automatically selected template; \textbf{(2) Thoracic segmentation:} the predicted mask resulted in partial or complete exclusion of the fetal heart; \textbf{(3) Slice-to-volume reconstruction:} the reconstructed cine volume did not reflect of the anatomy visible in the template stack; \textbf{(4) Anatomical reorientation:} the reoriented volume deviated substantially from the canonical fetal coordinate system. 

The automated pipeline was considered successful if no failure was identified. Each failure prompted a manual intervention at the corresponding stage, and the pipeline resumed from that point.

\subsection{Implementation details}

\subsubsection{Deep learning models}

All deep learning architectures were implemented in PyTorch using the MONAI framework \citep{paszke_pytorch_2019, cardoso_monai_2022} and trained on an NVIDIA RTX A5000 GPU with 24\,GB of memory. For thoracic and cardiac segmentation, a standard 3D U-Net architecture \citep{ronneberger_u-net_2015, milletari_v-net_2016} was employed, comprising $3\times3\times3$ convolutional layers, batch normalisation \citep{ioffe_batch_2015} and ReLU activations, in both encoder and decoder paths. The encoder path comprised strided convolutions to downsample the data. The decoder path used strided transpose convolutions to upscale the output to its original spatial dimensions. The number of features doubled at each encoding stage, resulting in a bottleneck consisting of 256 channels. For reorientation we employed a standard ViT architecture comprising $12$ transformer blocks, each with $12$ self-attention heads. The input volume was partitioned into $16 \times 16 \times 16$ patches and augmented with fixed sinusoidal positional embeddings \citep{dosovitskiy_image_2020}.

Hyperparameters (e.g.\ loss weighting coefficients, batch size, number of training epochs, and learning rate) were selected based on performance on the validation set. With the exception of the number of training epochs, all hyperparameters were kept identical across the four training strategies (source-only, target-only, joint, and transfer learning). The batch size and learning rate were set to 64 and $10^{-3}$ for the segmentation networks, and to 1024 and $10^{-4}$ for the reorientation network. For the reorientation network, the effective batch size exceeded the number of available training images. To compensate, each image was sampled multiple times per batch using independently drawn random rigid transformations, ensuring dense coverage of $\mathrm{SE}(3)$ at each gradient update. This sampling strategy was found to be critical for stabilising optimisation, particularly given the limited size of the target domain training set. The number of training epochs was set to $200$ for thoracic segmentation, $400$ for cardiac segmentation, and $2500$ for reorientation under the source-only and joint training configurations; these values were doubled for the target-only and transfer learning configurations to account for the smaller target domain training set. The model checkpoint corresponding to the lowest validation loss was retained at the end of training.

Prior to all training and testing experiments, images were corrected for bias-field inhomogeneity \citep{tustison_n4itk_2010} and resampled to isotropic $2\times2\times2$\,mm$^3$ resolution. Segmentation inference was performed using a sliding-window strategy \citep{cardoso_monai_2022}. Code for the segmentation and anatomical reorientation components is publicly available at \href{https://github.com/aboutill/SPARC}{https://github.com/aboutill/SPARC}.

\subsubsection{Slice-to-volume reconstruction}
\label{sec:software_recon}

The proposed reconstruction algorithm was implemented in C++ using a multi-threaded design, with image representation and registration handled via the MIRTK library\footnote{\href{https://mirtk.github.io/}{https://mirtk.github.io/}}. The method of \citet{van_amerom_fetal_2018} was obtained from SVRTK repository\footnote{\href{https://github.com/SVRTK/SVRTK}{https://github.com/SVRTK/SVRTK}}. Experiments were performed on a system equipped with 62\,GB of RAM, a 2.40\,GHz CPU, and 24 threads. For both methods, three interleaved registration-reconstruction cycles were performed, with the number of super-resolution reconstruction iterations set to 7, 7, and 21 in successive cycles respectively. Cine volumes were reconstructed at an isotropic spatial resolution of $1\times1\times1$\,mm$^3$. The code is publicly available at \href{https://github.com/baby-MedIA/svr-lite}{https://github.com/baby-MedIA/svr-lite}.

\subsubsection{Pipeline}

End-to-end pipeline experiments were conducted in semi-automatic mode on a system equipped with 96\,GB of RAM, a 2.40\,GHz CPU, and 24 threads. Timing was also recorded for the full end-to-end pipeline as well as for the reconstruction step. The \textbf{SPARC} pipeline is available in as a Docker container at \href{https://hub.docker.com/r/aboutill/sparc}{https://hub.docker.com/r/aboutill/sparc}. 

\section{Results}
\subsection{Thoracic segmentation}

\begin{table*}[ht!]
\centering
\begin{tabular}{| c | c | c | c | c | c |}
\hline
\multicolumn{2}{|c|}{Domain} & \multirow{2}{*}{Ensemble} & \multirow{2}{*}{$\textrm{DSC}$ [\%] $\uparrow$} & \multirow{2}{*}{$\textrm{HD}_{95\%}$ [mm] $\downarrow$} & \multirow{2}{*}{$\textrm{ASSD}$ [mm] $\downarrow$} \\ \cline{1-2}
Test & Training & & & & \\ \hline\hline
\multirow{2}{*}{Source} & Source & \ding{51} & $84.9\pm7.0$ & $4.1\pm3.0$ & $1.1\pm0.7$ \\ \cline{2-6}
& Joint & \ding{51} & $\mathbf{85.7\pm6.0}$ & $4.1\pm3.2$ & $1.1\pm0.7$ \\ \hline\hline
\multirow{6}{*}{Target} & Source & \ding{51} & $79.2\pm7.3$ & $10.8\pm10.6$ & $2.1\pm2.1$ \\ \cline{2-6}
& Target & \ding{51} & $81.8\pm4.5$ & $6.5\pm2.8$ & $1.4\pm0.4$ \\ \cline{2-6}
& Joint & \ding{51} & $84.3\pm3.9$ & $5.9\pm2.8$ & $1.2\pm0.4$ \\ \cline{2-6}
& \multirow{2}{*}{Transfer} & \ding{55} & $83.2\pm4.4$ & $6.9\pm4.4$ & $1.4\pm0.8$ \\ \cline{3-6}
& & \ding{51} & $\mathbf{84.7\pm3.9}$  & $\mathbf{5.4\pm2.4}$  & $\mathbf{1.1\pm0.4}$ \\ \cline{2-6}
& \multicolumn{2}{|c|}{Inter-rater} & $81.4\pm7.7$ & $6.6\pm4.0$ & $1.5\pm0.9$ \\ \hline
\end{tabular}
\caption{Quantitative evaluation of thoracic segmentation models on held-out test sets from the source (free-running; $m = 92$ stacks across $n = 23$ subjects) and target (Doppler-gated; $m = 36$ stacks across $n =5$ subjects) domains shown as mean$\pm$standard deviation. The best performance within each domain is shown in bold (linear mixed-effect model with Bonferroni correction, $p<0.05$).}
\label{table:results_chest_seg}
\end{table*}

Thoracic segmentation results are summarised in Table~\ref{table:results_chest_seg}. The transfer learning approach achieved the best performance on the target domain across all three metrics (DSC $84.7 \pm 3.9\%$, $\mathrm{HD}_{95\%}$ $5.4 \pm 2.4$\,mm, and ASSD $1.1 \pm 0.4$\,mm), with statistically significant improvements over all other approaches ($p < 0.05$). Notably, this performance is comparable the source-only model evaluated on the source domain (DSC $84.9 \pm 7.0\%$, $\mathrm{HD}_{95\%}$ $4.1 \pm 3.0$\,mm, ASSD $1.1 \pm 0.7$\,mm), and exceeds inter-rater agreement metrics (DSC $81.4 \pm 7.7\%$, $\mathrm{HD}_{95\%}$ $6.6 \pm 4.0$\,mm, ASSD $1.5 \pm 0.9$\,mm).

The substantially degraded performance of the source-only model on the target domain ($\mathrm{HD}_{95\%}$ $10.8 \pm 10.6$\,mm, DSC $79.2 \pm 7.3\%$) confirms a need for our domain adaptation strategy. Figure~\ref{fig:thorax_seg} provides a qualitative comparison of the four training strategies on three representative target domain subjects, illustrating the superior boundary delineation of the Transfer approach.

\subsection{Reconstruction}

\begin{table}[ht!]
\centering%
\resizebox{\columnwidth}{!}{\begin{tabular}{| c | c | c | c |}
\hline
\multicolumn{2}{|c|}{Method} & van Amerom & Proposed \\ \hline

\multirow{2}{*}{\shortstack{Model\\assumptions}} & Motion states & Frame-wise & Slice-wise \\ \cline{2-4}
& tPSF & \ding{51} & \ding{55} \\ \hline

\multirow{4}{*}{\shortstack{Step\\complexity}} & \multirow{2}{*}{\shortstack{Slice/frame-to-volume\\registration}} & \multirow{2}{*}{$\bigO{(KT)}$} & \multirow{2}{*}{$\bigO{(K)}$} \\ 
& & & \\ \cline{2-4}
& \multirow{2}{*}{\shortstack{Super-resolution\\reconstruction}} & \multirow{2}{*}{$\bigO{(NT^2)}$} & \multirow{2}{*}{$\bigO{(NT)}$} \\ 
& & & \\ \hline

\multirow{2}{*}{\shortstack{Leave-stack\\-out}} 
& NCC $\uparrow$ & $0.89 \pm 0.04$ & $\mathbf{0.91 \pm 0.03}$ \\  \cline{2-4} %
& NRMSE $\downarrow$ & $0.13 \pm 0.03$ & $\mathbf{0.11 \pm 0.02}$ \\ \hline %

\multirow{2}{*}{\shortstack{Data\\consistency}} 
& NCC $\uparrow$ & $\mathbf{0.97 \pm 0.01}$ & $0.96 \pm 0.01$ \\ \cline{2-4} %
& NRMSE $\downarrow$ & $\mathbf{0.09 \pm 0.01}$ & $0.10 \pm 0.02$ \\ \hline %

\multirow{2}{*}{\shortstack{Slice/frame\\ratios}} 
& Included [$\%$] $\uparrow$ & $55.0 \pm 5.7$ & $\mathbf{78.2 \pm 5.9}$ \\ \cline{2-4} %
& Excluded [$\%$] $\downarrow$ & $45.0 \pm 5.7$ & $\mathbf{21.8 \pm 5.9}$ \\ \hline  %
\multicolumn{2}{|c|}{Time [min] $\downarrow$} & $49.0 \pm 14.1$ & $\mathbf{4.8 \pm 1.0}$ \\ \hline 

\end{tabular}}
\caption{Quantitative evaluation of the reconstruction algorithms on the target domain test set (Doppler-gated; $n = 10$ subjects) shown as mean$\pm$standard deviation. The best performance within each domain is shown in bold (paired $t$-test, $p < 0.05$)}
\label{table:results_recon}
\end{table}

\begin{figure*}[t]
\centering
\begin{adjustbox}{width=\textwidth}
\begin{tikzpicture}[every node/.style={inner sep=0,outer sep=0}]

\node[inner sep=0pt] at (0,0)
    {\includegraphics[height=.048\textwidth,trim=10 15 10 5, clip]{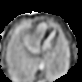}};
\node[inner sep=0pt] at (0,-.85)
    {\includegraphics[height=.048\textwidth,trim=10 15 10 5, clip]{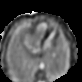}};

\node[inner sep=0pt] at (.84,0)
    {\includegraphics[height=.048\textwidth,trim=5 5 15 5, clip]{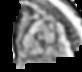}};
\node[inner sep=0pt] at (.84,-.85)
    {\includegraphics[height=.048\textwidth,trim=5 5 15 5, clip]{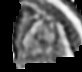}};

\node[inner sep=0pt] at (1.66,0)
    {\includegraphics[height=.048\textwidth,trim=12 5 17 15, clip]{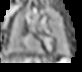}};
\node[inner sep=0pt] at (1.66,-.85)
    {\includegraphics[height=.048\textwidth,trim=12 5 17 15, clip]{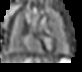}};

\node[inner sep=0pt] at (2.48,0)
    {\includegraphics[width=.048\textwidth,height=.045\textwidth,angle=-90,trim=15 0 35 0, clip]{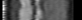}};
\node[inner sep=0pt] at (2.48,-.85)
    {\includegraphics[width=.048\textwidth,height=.045\textwidth,angle=-90,trim=15 0 35 0, clip]{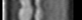}};

\node[inner sep=0pt] at (3.5,0)
    {\includegraphics[height=.048\textwidth,trim=15 15 5 5, clip]{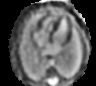}};
\node[inner sep=0pt] at (3.5,-.85)
    {\includegraphics[height=.048\textwidth,trim=15 15 5 5, clip]{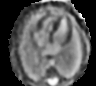}};

\node[inner sep=0pt] at (4.48,0)
    {\includegraphics[height=.048\textwidth,trim=5 5 15 5, clip]{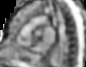}};
\node[inner sep=0pt] at (4.48,-.85)
    {\includegraphics[height=.048\textwidth,trim=5 5 15 5, clip]{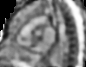}};

\node[inner sep=0pt] at (5.41,0)
    {\includegraphics[height=.048\textwidth,trim=20 5 10 5, clip]{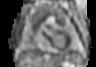}};
\node[inner sep=0pt] at (5.41,-.85)
    {\includegraphics[height=.048\textwidth,trim=20 5 10 5, clip]{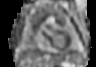}};

\node[inner sep=0pt] at (6.27,0)
    {\includegraphics[width=.048\textwidth,height=.045\textwidth,angle=90,trim=12 0 36 0, clip]{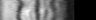}};
\node[inner sep=0pt] at (6.27,-.85)
    {\includegraphics[width=.048\textwidth,height=.045\textwidth,angle=90,trim=12 0 36 0, clip]{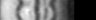}};

\node[inner sep=0pt,rotate=90] at (-.5,0) {\scalebox{.35}{van Amerom et al.}};
\node[inner sep=0pt,rotate=90] at (-.5,-.85) {\scalebox{.35}{Proposed}};

\node[inner sep=0pt] at (1.26,.65) {\scalebox{.35}{Case 1}};
\node[inner sep=0pt] at (0,.5)    {\scalebox{.35}{Axial}};
\node[inner sep=0pt] at (.84,.5)  {\scalebox{.35}{Sagittal}};
\node[inner sep=0pt] at (1.68,.5) {\scalebox{.35}{Coronal}};
\node[inner sep=0pt] at (2.48,.5) {\scalebox{.35}{Temporal}};

\node[inner sep=0pt] at (4.945,.65) {\scalebox{.35}{Case 2}};
\node[inner sep=0pt] at (3.5,.5)  {\scalebox{.35}{Axial}};
\node[inner sep=0pt] at (4.48,.5) {\scalebox{.35}{Sagittal}};
\node[inner sep=0pt] at (5.41,.5) {\scalebox{.35}{Coronal}};
\node[inner sep=0pt] at (6.27,.5) {\scalebox{.35}{Temporal}};

\draw[line width=0.1mm, color=green] (-.1,.15) -- (.28,-.05);
\draw[line width=0.1mm, color=green] (-.1,-.7) -- (.28,-.9);

\draw[line width=0.1mm, color=green] (3.28,.22) -- (3.74,.11);
\draw[line width=0.1mm, color=green] (3.28,-.63) -- (3.74,-.74);

\draw[line width=0.1mm, color=blue, -{Latex[length=1mm, width=1mm]}] (2.55,.20) -- (2.4,-.05);
\draw[line width=0.1mm, color=blue, -{Latex[length=1mm, width=1mm]}] (2.55,-.65) -- (2.4,-.9);

\end{tikzpicture}
\end{adjustbox}
\caption{Qualitative comparison of cine volumes reconstructed by \citet{van_amerom_fetal_2019} and our proposed methods, for two representative target domain cases from the leave-stack-out experiment. Three orthogonal views are shown, alongside a temporal intensity profile computed along a fixed line (\textcolor{green}{\raisebox{1.8pt}{\rule{5pt}{1pt}}}) across all $T = 20$ cardiac phases.}
\label{fig:recon}
\end{figure*}
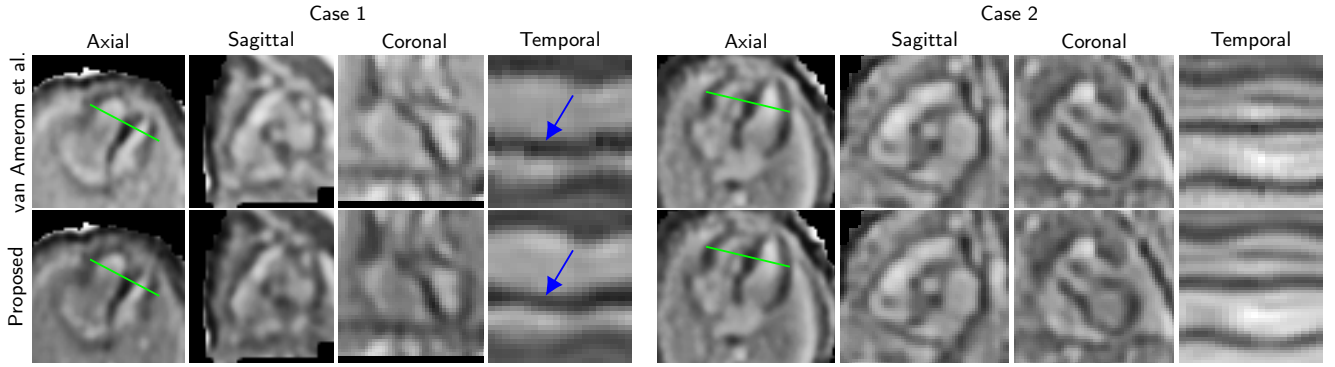

Reconstruction results are summarised in Table~\ref{table:results_recon}. The proposed method achieved significantly higher leave-stack-out NCC ($0.91 \pm 0.03$ vs $0.89 \pm 0.04$) and lower NRMSE ($0.11 \pm 0.02$ vs $0.13 \pm 0.03$), indicating superior generalisation to unseen data and more faithful recovery of the underlying anatomy. This shows that assuming a shared motion state per slice rather than per frame reduces the risk of overfitting to the input data. Marginally higher consistency combined with smaller fraction of included input data in the method of \citet{van_amerom_fetal_2019} points to overfitting rather then genuinely better performance compared to our method. In addition, the proposed method achieved a tenfold reduction in reconstruction time ($4.8 \pm 1.0$ vs $49.0 \pm 14.1$ min, $p < 0.05$), directly attributable to the reduced computational complexity: from $\mathcal{O}(KT)$ to $\mathcal{O}(K)$ for slice-to-volume registration, and from $\mathcal{O}(NT^2)$ to $\mathcal{O}(NT)$ for super-resolution reconstruction (with $T = 20$). Figure~\ref{fig:recon} demonstrates that our method produces notably smoother temporal transitions in the cardiac phase profile than \citet{van_amerom_fetal_2019}, with comparable spatial sharpness. 

\subsection{Cardiac segmentation}

\begin{table*}[t]
\centering
\begin{tabular}{| c | c | c | c | c | c | c |}
\hline
\multicolumn{2}{|c|}{Domain} & \multirow{2}{*}{Ensemble} & \multirow{2}{*}{$\textrm{DSC}$ [\%] $\uparrow$} & \multirow{2}{*}{$\textrm{HD}_{95\%}$ [mm] $\downarrow$} & \multirow{2}{*}{$\textrm{ASSD}$ [mm] $\downarrow$} & \multirow{2}{*}{$\textrm{CD}$ [mm] $\downarrow$} \\ \cline{1-2}
Test & Training & & & & & \\ \hline\hline
\multirow{2}{*}{Source} & Source & \ding{51} & $83.4\pm4.9$ & $4.8\pm1.9$ & $1.7\pm0.4$ & $4.2 \pm 2.3$\\ \cline{2-7}
& Joint & \ding{51} & $83.8\pm5.0$ & $4.7\pm1.5$ & $1.6\pm0.4$ & $4.0 \pm 2.3$ \\ \hline\hline
\multirow{5}{*}{Target} & Source & \ding{51} & $20.0\pm30.8$ & $27.8\pm12.0$ & $14.4\pm7.7$ & $39.5 \pm 22.3$ \\ \cline{2-7}
& Target & \ding{51} & $73.8\pm9.0$ & $8.2\pm3.6$ & $3.2\pm1.4$ & $12.3 \pm 5.2$ \\ \cline{2-7}
& Joint & \ding{51} & $71.0\pm7.1$ & $9.2\pm1.4$ & $3.5\pm0.9$ & $11.1 \pm 7.6$ \\ \cline{2-7}
& \multirow{2}{*}{Transfer} & \ding{55} & $79.6\pm2.5$ & $6.2\pm1.3$ & $2.4\pm0.5$ & $8.4 \pm 2.1$ \\ \cline{3-7}
& & \ding{51} & $82.2\pm2.1$  & $5.1\pm0.6$  & $2.1\pm0.4$ & $6.3 \pm 2.9$ \\ \hline
\end{tabular}
\caption{Quantitative evaluation of cardiac segmentation models on held-out test sets from the source (free-running; $n = 22$ subjects) and target (Doppler-gated; $n =5$ subjects) domains shown as mean$\pm$standard deviation.}
\label{table:results_seg_heart}
\end{table*}

Cardiac segmentation results are summarised in Table~\ref{table:results_seg_heart}. The transfer learning approach with ensemble achieved the best performance on the target domain across all metrics (DSC $82.2 \pm 2.1\%$, $\mathrm{HD}_{95\%}$ $5.1 \pm 0.6$\,mm, ASSD $2.1 \pm 0.4$\,mm, CD $6.3 \pm 2.9$\,mm), reaching performance comparable to that of the source-only model on the source domain (DSC $83.4 \pm 4.9\%$, $\mathrm{HD}_{95\%}$ $4.8 \pm 1.9$\,mm, ASSD $1.7 \pm 0.4$\,mm, CD $4.2 \pm 2.3$\,mm), while significantly outperforming target-only (DSC $73.8 \pm 9.0\%$) and source-only (DSC $20.0 \pm 30.8\%$) model. These findings are consistent with the results for thoracic segmentation. 

The relatively low centre distance achieved by the proposed approach demonstrates that cardiac segmentation provides a robust spatial initialisation for the subsequent reorientation network, as further confirmed by the ablation study (Section~\ref{sec:results_pipeline}).

\subsection{Reorientation}
\label{sec:results_reorientation}

\begin{table}[t!]
\centering
\resizebox{\columnwidth}{!}{
\begin{tabular}{| c | c | c | c | c |}
\hline
\multicolumn{2}{|c|}{Domain} & \multirow{2}{*}{Ensemble} & \multirow{2}{*}{$\textrm{GD}$ [$\degree$] $\downarrow$} & \multirow{2}{*}{$\textrm{CD}$ [mm] $\downarrow$} \\ \cline{1-2}
Test & Training & & & \\ \hline\hline
\multirow{2}{*}{Source} & Source & Chordal & $13.1\pm6.3$ & $1.6\pm1.5$ \\ \cline{2-5}
& Joint & Chordal & $\mathbf{10.5\pm5.8}$ & $\mathbf{1.1\pm1.0}$ \\ \hline\hline
\multirow{8}{*}{Target} & Source & Chordal & $74.9\pm52.8$ & $5.2\pm4.5$ \\ \cline{2-5}
& Target & Chordal & $33.4\pm36.4$ & $4.4\pm4.5$  \\ \cline{2-5}
& Joint & Chordal & $16.0\pm11.8$ & $2.9\pm2.5$ \\ \cline{2-5}
& \multirow{4}{*}{Transfer} & \ding{55} & $18.9\pm13.5$ & $6.9\pm3.9$ \\ \cline{3-5}
& & Quaternion & $16.7\pm15.3$ & $2.4\pm1.9$ \\ \cline{3-5}
& & Geodesic & $16.3\pm13.5$ & $2.4\pm1.9$ \\ \cline{3-5}
& & Chordal & $\mathbf{14.3\pm10.9}$  & $\mathbf{2.4\pm1.9}$ \\ \cline{2-5}
& \multicolumn{2}{|c|}{Inter-rater\textsuperscript{\textdagger}} & $7.7\pm2.9$  & \---  \\ \hline
\end{tabular}}
\caption{Quantitative evaluation of reorientation models on held-out test sets from the source (free-running; $n =21$ subjects) and target (Doppler-gated; $n =7$ subjects) domains shown as mean$\pm$standard deviation. Each image was sampled $250$ times using independently drawn random rigid transformations. The best performance within each domain is shown in bold (linear mixed-effect model comparison with Bonferroni correction, $p<0.05$). \textsuperscript{\textdagger}Target domain GD inter-rater agreement was computed on a single reorientation per subject ($n = 7$) and is not directly comparable to network performance assessed over whole range of rotations.}

\label{table:results_reorientation}
\end{table}

Reorientation results are summarised in Table~\ref{table:results_reorientation}. The transfer-learning approach with chordal averaging achieved statistically superior performance on the target domain across all metrics (GD $14.3 \pm 10.9\degree$, CD $2.4 \pm 1.9$\,mm, $p < 0.05$), and reached performance comparable to that of the source-only model evaluated on the source domain (GD $13.1 \pm 6.3\degree$, CD $1.6 \pm 1.5$\,mm). The source-only model fails on the target domain (GD $74.9 \pm 52.8\degree$), while the comparatively poor performance of the target-only model (GD $33.4 \pm 36.4\degree$) reflects the limited size of the target domain training set.
All three ensemble strategies (quaternion, geodesic, and chordal averaging \citep{hartley_rotation_2013}) provided consistent improvements over the mean performance of individual networks. The best performing chordal averaging was selected and the ensemble method of choice.

\begin{figure*}[t!]
\centering
\begin{adjustbox}{width=\textwidth}
\begin{tikzpicture}[every node/.style={inner sep=0,outer sep=0}]

\node[inner sep=0pt] at (0,0)
    {\includegraphics[width=.048\textwidth, trim=5 5 5 5, clip]{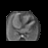}};
\node[inner sep=0pt] at (0,-.85)
    {\includegraphics[width=.048\textwidth, trim=5 5 5 5, clip]{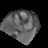}};
\node[inner sep=0pt] at (0,-1.7) 
    {\includegraphics[width=.048\textwidth, trim=5 5 5 5, clip]{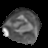}};
\node[inner sep=0pt] at (0,-2.55)
    {\includegraphics[width=.048\textwidth, trim=5 5 5 5, clip]{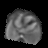}};
\node[inner sep=0pt] at (0,-3.4)
    {\includegraphics[width=.048\textwidth, trim=5 5 5 5, clip]{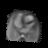}};
\node[inner sep=0pt] at (0,-4.25) 
    {\includegraphics[width=.048\textwidth, trim=5 5 5 5, clip]{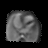}};

\node[inner sep=0pt] at (.85,0)
    {\includegraphics[width=.048\textwidth, trim=5 5 5 5, clip]{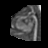}};
\node[inner sep=0pt] at (.85,-.85)
    {\includegraphics[width=.048\textwidth, trim=5 5 5 5, clip]{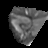}};
\node[inner sep=0pt] at (.85,-1.7) 
    {\includegraphics[width=.048\textwidth, trim=5 5 5 5, clip]{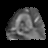}};
\node[inner sep=0pt] at (.85,-2.55)
    {\includegraphics[width=.048\textwidth, trim=5 5 5 5, clip]{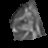}};
\node[inner sep=0pt] at (.85,-3.4)
    {\includegraphics[width=.048\textwidth, trim=5 5 5 5, clip]{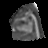}};
\node[inner sep=0pt] at (.85,-4.25) 
    {\includegraphics[width=.048\textwidth, trim=5 5 5 5, clip]{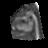}};

\node[inner sep=0pt] at (1.7,0)
    {\includegraphics[width=.048\textwidth, trim=5 5 5 5, clip]{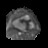}};
\node[inner sep=0pt] at (1.7,-.85)
    {\includegraphics[width=.048\textwidth, trim=5 5 5 5, clip]{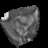}};
\node[inner sep=0pt] at (1.7,-1.7) 
    {\includegraphics[width=.048\textwidth, trim=5 5 5 5, clip]{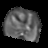}};
\node[inner sep=0pt] at (1.7,-2.55)
    {\includegraphics[width=.048\textwidth, trim=5 5 5 5, clip]{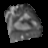}};
\node[inner sep=0pt] at (1.7,-3.4)
    {\includegraphics[width=.048\textwidth, trim=5 5 5 5, clip]{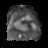}};
\node[inner sep=0pt] at (1.7,-4.25) 
    {\includegraphics[width=.048\textwidth, trim=5 5 5 5, clip]{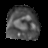}};

\node[inner sep=0pt] at (2.65,0)
    {\includegraphics[width=.048\textwidth, trim=5 5 5 5, clip]{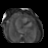}};
\node[inner sep=0pt] at (2.65,-.85)
    {\includegraphics[width=.048\textwidth, trim=5 5 5 5, clip]{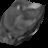}};
\node[inner sep=0pt] at (2.65,-1.7) 
    {\includegraphics[width=.048\textwidth, trim=5 5 5 5, clip]{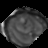}};
\node[inner sep=0pt] at (2.65,-2.55)
    {\includegraphics[width=.048\textwidth, trim=5 5 5 5, clip]{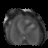}};
    \node[inner sep=0pt] at (2.65,-3.4)
    {\includegraphics[width=.048\textwidth, trim=5 5 5 5, clip]{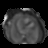}};
\node[inner sep=0pt] at (2.65,-4.25) 
    {\includegraphics[width=.048\textwidth, trim=5 5 5 5, clip]{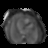}};

\node[inner sep=0pt] at (3.5,0)
    {\includegraphics[width=.048\textwidth, trim=5 5 5 5, clip]{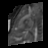}};
\node[inner sep=0pt] at (3.5,-.85)
    {\includegraphics[width=.048\textwidth, trim=5 5 5 5, clip]{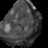}};
\node[inner sep=0pt] at (3.5,-1.7) 
    {\includegraphics[width=.048\textwidth, trim=5 5 5 5, clip]{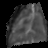}};
\node[inner sep=0pt] at (3.5,-2.55)
    {\includegraphics[width=.048\textwidth, trim=5 5 5 5, clip]{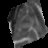}};
\node[inner sep=0pt] at (3.5,-3.4)
    {\includegraphics[width=.048\textwidth, trim=5 5 5 5, clip]{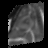}};
\node[inner sep=0pt] at (3.5,-4.25) 
    {\includegraphics[width=.048\textwidth, trim=5 5 5 5, clip]{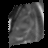}};

\node[inner sep=0pt] at (4.35,0)
    {\includegraphics[width=.048\textwidth, trim=5 5 5 5, clip]{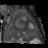}};
\node[inner sep=0pt] at (4.35,-.85)
    {\includegraphics[width=.048\textwidth, trim=5 5 5 5, clip]{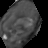}};
\node[inner sep=0pt] at (4.35,-1.7) 
    {\includegraphics[width=.048\textwidth, trim=5 5 5 5, clip]{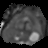}};
\node[inner sep=0pt] at (4.35,-2.55)
    {\includegraphics[width=.048\textwidth, trim=5 5 5 5, clip]{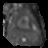}};
\node[inner sep=0pt] at (4.35,-3.4)
    {\includegraphics[width=.048\textwidth, trim=5 5 5 5, clip]{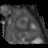}};
\node[inner sep=0pt] at (4.35,-4.25) 
    {\includegraphics[width=.048\textwidth, trim=5 5 5 5, clip]{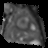}};

\node[inner sep=0pt] at (5.3,0)
    {\includegraphics[width=.048\textwidth, trim=5 5 5 5, clip]{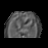}};
\node[inner sep=0pt] at (5.3,-.85)
    {\includegraphics[width=.048\textwidth, trim=5 5 5 5, clip]{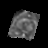}};
\node[inner sep=0pt] at (5.3,-1.7) 
    {\includegraphics[width=.048\textwidth, trim=5 5 5 5, clip]{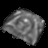}};
\node[inner sep=0pt] at (5.3,-2.55)
    {\includegraphics[width=.048\textwidth, trim=5 5 5 5, clip]{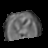}};
    \node[inner sep=0pt] at (5.3,-3.4)
    {\includegraphics[width=.048\textwidth, trim=5 5 5 5, clip]{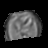}};
\node[inner sep=0pt] at (5.3,-4.25) 
    {\includegraphics[width=.048\textwidth, trim=5 5 5 5, clip]{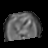}};

\node[inner sep=0pt] at (6.15,0)
    {\includegraphics[width=.048\textwidth, trim=5 5 5 5, clip]{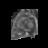}};
\node[inner sep=0pt] at (6.15,-.85)
    {\includegraphics[width=.048\textwidth, trim=5 5 5 5, clip]{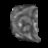}};
\node[inner sep=0pt] at (6.15,-1.7) 
    {\includegraphics[width=.048\textwidth, trim=5 5 5 5, clip]{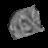}};
\node[inner sep=0pt] at (6.15,-2.55)
    {\includegraphics[width=.048\textwidth, trim=5 5 5 5, clip]{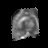}};
\node[inner sep=0pt] at (6.15,-3.4)
    {\includegraphics[width=.048\textwidth, trim=5 5 5 5, clip]{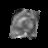}};
\node[inner sep=0pt] at (6.15,-4.25) 
    {\includegraphics[width=.048\textwidth, trim=5 5 5 5, clip]{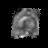}};

\node[inner sep=0pt] at (7,0)
    {\includegraphics[width=.048\textwidth, trim=5 5 5 5, clip]{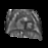}};
\node[inner sep=0pt] at (7,-.85)
    {\includegraphics[width=.048\textwidth, trim=5 5 5 5, clip]{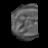}};
\node[inner sep=0pt] at (7,-1.7) 
    {\includegraphics[width=.048\textwidth, trim=5 5 5 5, clip]{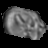}};
\node[inner sep=0pt] at (7,-2.55)
    {\includegraphics[width=.048\textwidth, trim=5 5 5 5, clip]{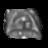}};
\node[inner sep=0pt] at (7,-3.4)
    {\includegraphics[width=.048\textwidth, trim=5 5 5 5, clip]{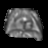}};
\node[inner sep=0pt] at (7,-4.25) 
    {\includegraphics[width=.048\textwidth, trim=5 5 5 5, clip]{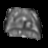}};

\node[inner sep=0pt,rotate=90] at (-.5,0) {\scalebox{.35}{Ground truth}};
\node[inner sep=0pt,rotate=90] at (-.5,-.85) {\scalebox{.35}{Simulated}};
\node[inner sep=0pt,rotate=90] at (-.5,-1.7) {\scalebox{.35}{Source-only}};
\node[inner sep=0pt,rotate=90] at (-.5,-2.55) {\scalebox{.35}{Target-only}};
\node[inner sep=0pt,rotate=90] at (-.5,-3.4) {\scalebox{.35}{Joint}};
\node[inner sep=0pt,rotate=90] at (-.5,-4.25) {\scalebox{.35}{Transfer}};

\node[inner sep=0pt] at (.85,.65) {\scalebox{.35}{Case 1}};
\node[inner sep=0pt] at (0,.5) {\scalebox{.35}{Axial}};
\node[inner sep=0pt] at (.85,.5) {\scalebox{.35}{Sagittal}};
\node[inner sep=0pt] at (1.7,.5) {\scalebox{.35}{Coronal}};

\node[inner sep=0pt] at (3.5,.65) {\scalebox{.35}{Case 2}};
\node[inner sep=0pt] at (2.65,.5) {\scalebox{.35}{Axial}};
\node[inner sep=0pt] at (3.5,.5) {\scalebox{.35}{Sagittal}};
\node[inner sep=0pt] at (4.35,.5) {\scalebox{.35}{Coronal}};

\node[inner sep=0pt] at (6.15,.65) {\scalebox{.35}{Case 3}};
\node[inner sep=0pt] at (5.3,.5) {\scalebox{.35}{Axial}};
\node[inner sep=0pt] at (6.15,.5) {\scalebox{.35}{Sagittal}};
\node[inner sep=0pt] at (7,.5) {\scalebox{.35}{Coronal}};

\end{tikzpicture}
\end{adjustbox}
\caption{Qualitative comparison of reorientation for three target domain subjects across the four training strategies (source-only, target-only, joint, and transfer). Cases were selected to represent varying levels of image contrast with distinct simulated random transformation. Time-averaged cine volumes are shown in canonical fetal coordinate system across three orthogonal views. Images resolution is set to the input resolution of the ViT reorientation network, $2 \times 2 \times 2$ mm.}
\label{fig:reorientation}
\end{figure*}
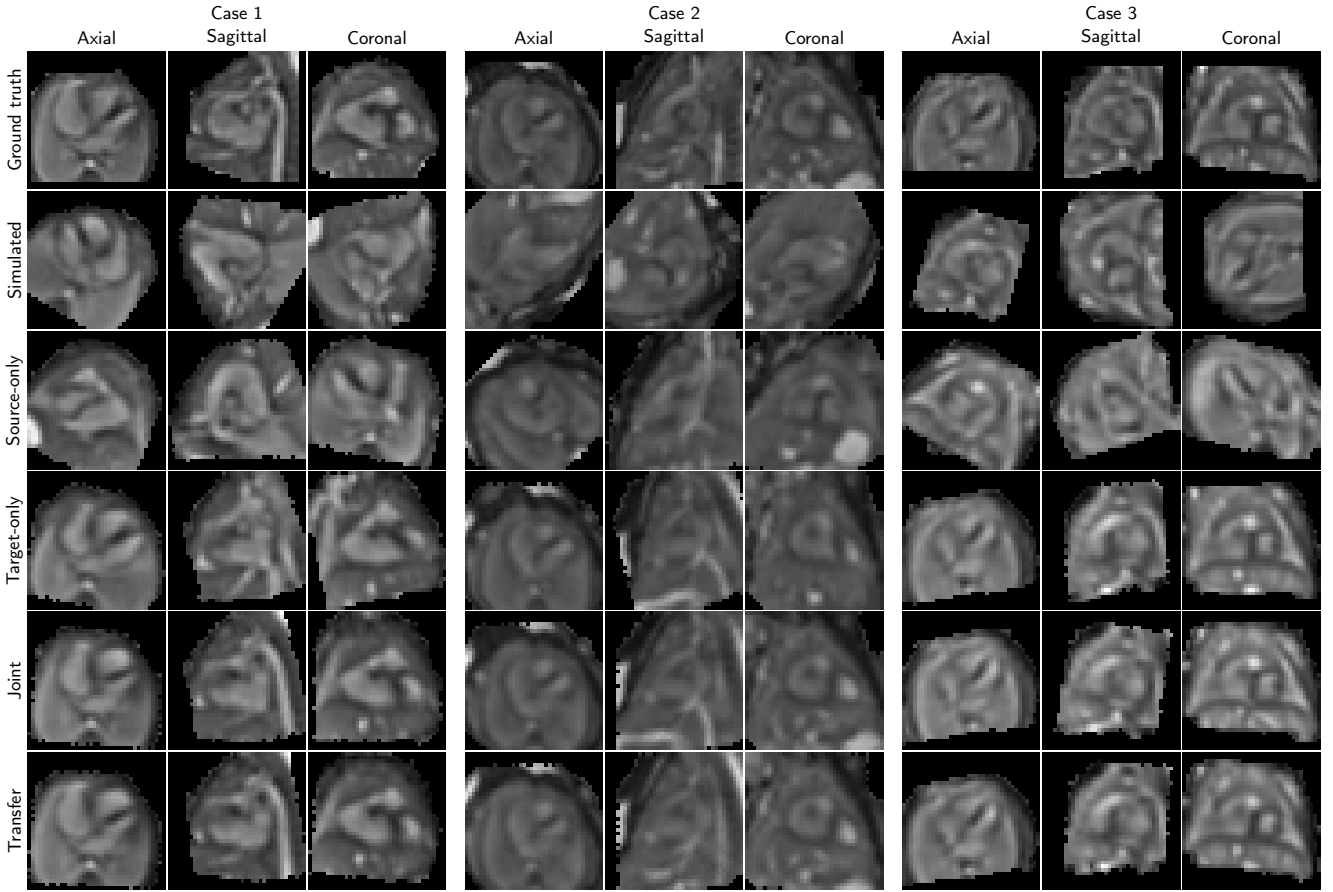

Figure~\ref{fig:reorientation} illustrates the superior visual agreement of the proposed approach with the ground truth reorientation across all cases and orientations. Furthermore, Figure~\ref{fig:reorientation_subject} reveals that GD remains low and approximately constant across the full range of simulated orientations, with the majority of error concentrated in a small number of challenging subjects. 

\subsection{End-to-end pipeline}
\label{sec:results_pipeline}

\begin{table}[t!]

\centering
\resizebox{\columnwidth}{!}{
\begin{tabular}{| c | c | c | c |}
\hline
\multicolumn{4}{|c|}{\textbf{\textbf{SPARC} pipeline evaluation on clinical cohort ($n = 121)$}} \\ \hline\hline

\multicolumn{4}{|c|}{\textbf{1) Manual intervention summary}}  \\ \hline
\multicolumn{2}{|c|}{Step} & $n$ ($\%$) & Time per case  \\ \hline
\multicolumn{2}{|c|}{Manual stack exclusion} & 6 (5.0\%) & $\sim$ 30 s  \\ \hline
\multicolumn{2}{|c|}{Manual template selection} & 3 (2.5\%) & $\sim$ 60 s  \\ \hline
\multicolumn{2}{|c|}{Manual thoracic segmentation} & 3 (2.5\%) & $\sim$ 120 s \\ \hline
\multicolumn{2}{|c|}{Manual anatomical reorientation} & 12 (9.9\%) & $\sim$ 60 s \\ \hline
\multicolumn{2}{|c|}{\textbf{Multiple interventions}} & \textbf{2 (1.7\%)} & \--- \\ \hline
\multicolumn{2}{|c|}{\textbf{Unique cases}} & \textbf{21 (17.4\%)} & \--- \\
\hline\hline

\multicolumn{4}{|c|}{\textbf{2) Template selection}} \\ \hline
\multicolumn{2}{|c|}{Template stack evaluation}  & \multicolumn{2}{|c|}{$n$ ($\%$)} \\ \hline
\multicolumn{2}{|c|}{Failure} & \multicolumn{2}{|c|}{$3$ ($2.5\%$)} \\ \hline
\multicolumn{2}{|c|}{Success} & \multicolumn{2}{|c|}{$118$ ($97.5\%$)} \\ \hline\hline

\multicolumn{4}{|c|}{\textbf{3) Thoracic segmentation}} \\ \hline
\multicolumn{2}{|c|}{Segmentation evaluation}  & \multicolumn{2}{|c|}{$n$ ($\%$)} \\ \hline
\multicolumn{2}{|c|}{Failure} & \multicolumn{2}{|c|}{$3$ ($2.5\%$)} \\ \hline
\multicolumn{2}{|c|}{Success} & \multicolumn{2}{|c|}{$118$ ($97.5\%$)} \\ \hline\hline

\multicolumn{4}{|c|}{\textbf{4) Slice-to-volume reconstruction}} \\ \hline
\multicolumn{2}{|c|}{Reconstruction evaluation}  & $n$ ($\%$) & Time [min] \\ \hline
\multicolumn{2}{|c|}{Failure} & \--- & \--- \\ \hline
\multicolumn{2}{|c|}{Success} & $121$ ($100\%$) & $4.9 \pm 1.0$ \\ \hline\hline

\multicolumn{4}{|c|}{\textbf{5) Anatomical reorientation}} \\ \hline
\multirow{2}{*}{\shortstack{Cardiac\\localization}} & \multirow{2}{*}{\shortstack{Reorientation\\evaluation}} & \multicolumn{2}{|c|}{\multirow{2}{*}{$n$ ($\%$)}} \\
& & \multicolumn{2}{|c|}{} \\ \hline
\multirow{2}{*}{\ding{55}} & Failure & \multicolumn{2}{|c|}{$31$ ($25.6\%$)} \\ \cline{2-4}
 & Success & \multicolumn{2}{|c|}{$90$ ($74.4\%$)} \\ \hline
\multirow{2}{*}{\ding{51}} & Failure & \multicolumn{2}{|c|}{$12$ ($9.9\%$)} \\ \cline{2-4}
 & Success & \multicolumn{2}{|c|}{$109$ ($90.1\%$)} \\ \hline\hline

\multicolumn{4}{|c|}{\textbf{6) End-to-end pipeline summary}} \\ \hline
\multicolumn{2}{|c|}{Pipeline evaluation} & $n$ ($\%$) & Time [min] \\\hline
\multicolumn{2}{|c|}{Semi-automatic} & 21 (17.4) & \---  \\ \hline
\multicolumn{2}{|c|}{\textbf{Fully-automatic}} & \textbf{100 (82.6)} & $7.1 \pm 1.3$ \\ \hline

\end{tabular}}
\caption{End-to-end \textbf{SPARC} pipeline evaluation on a held-out clinical cohort ($n = 121$ subjects). \textbf{1)} Manual intervention summary: the number and estimated duration of manual interventions required at each pipeline stage. \textbf{2)} Template stack selection evaluation. \textbf{3)} Thoracic segmentation evaluation. \textbf{4)} Slice-to-volume reconstruction evaluation with mean processing time. \textbf{5)} Anatomical reorientation evaluation stratified by cardiac centre localisation ablation. \textbf{6)} End-to-end pipeline summary: overall counts and mean processing time for fully automatic and semi-automatic runs.}
\label{table:results_pipeline}
\end{table}

The results of the end-to-end pipeline evaluation are summarised in Table~\ref{table:results_pipeline}. Prior to end-to-end pipeline evaluation, input stacks were manually excluded in $n = 6$ cases due to excessive within-stack motion corruption. Both automatic template selection and thoracic segmentation was successful in $118$ out of $121$ cases ($97.5\%$). Slice-to-volume reconstruction was successful in all $121$ cases. Mean reconstruction time was $4.9 \pm 1.0$ min per case. The cardiac centre localisation step was critical in reducing the reorientation failure rate from $25.6\%$ ($31/121$) to $9.9\%$ ($12/121$).

The complete \textbf{SPARC} pipeline ran fully automatically in $100$ out of $121$ cases ($82.6\%$), with a mean end-to-end processing time of $7.1 \pm 1.3$ min per case. With manual intervention, all subjects whose data quality passed the inclusion criteria were successfully reconstructed.

\section{Discussion}

This study introduces an end-to-end, deep learning-enabled pipeline for the acquisition and reconstruction of 3D$+$time fetal cardiac MRI. Leveraging a recently introduced, CE-marked Doppler ultrasound gating device \citep{kording_evaluation_2018, kording_dynamic_2018}, the proposed framework provides a solution deployable in routine clinical practice and large-scale research studies.  By reducing fully automated reconstruction times to minutes while offering efficient manual intervention options, the pipeline ensures robust processing for cases meeting minimal inclusion criteria (at least four stacks without major motion corruption and successful Doppler gating).

\subsection{Clinical translation}
\label{sec:clinical_translation}

The \textbf{SPARC} pipeline is currently deployed at our institution as part of fetal cardiac MRI screening. The Doppler ultrasound gating device used in this study has already been adopted at multiple fetal cardiac MRI centres \citep{ tavares_de_sousa_dynamic_2019, vollbrecht_fetal_2024}, providing a natural pathway for deployment beyond our institution. The reconstruction algorithm requires no site-specific adaptation and can be used out-of-the-box. While the deep learning components are susceptible to domain shifts, our proposed transfer learning approach allows efficient retraining with a small amount of locally acquired and annotated data, using our publicly available model weights and training code. 

Additionally, the competitive results of joint training strategy provide evidence that pooling heterogeneous multi-site data is a viable generalisation strategy \citep{wang_generalizing_2023}. The public availability of the model weights and training code also supports federated learning \citep{guan_federated_2024}, as institutions can initialise from the provided weights and contribute site-specific fine-tuning without exposing patient data. In parallel, the emergence of foundation models pre-trained on large and diverse medical imaging datasets may reduce the annotation burden at new sites \citep{chen_foundation_2026}, either as initialisations for fine-tuning or as zero-shot segmentation tools.

Automated quality control could further enhance deployability to clinical practice by automatically highlighting reconstructions that require further inspection. Our current implementation makes use of agreements between our ensemble deep learning models and consistency metrics in super-resolution reconstruction (see Figure~\ref{fig:pipeline_qc}). By identifying suitable thresholds for these QC metrics, we can establish safe operating zones where the fully automated pipeline can be trusted, thus significantly reducing the workload of clinicians. 

\subsection{Limitations}

Our pipeline relies on successful Doppler gating, which can be lost with extreme fetal motion. Subsequent repositioning of the device results in temporal misalignment of the stacks. Automatic detection and re-alignment of temporally misaligned stacks \citep{jansz_metric_2010} represents a natural direction for future work.

We adopted a rigid motion model under the assumption of limited anatomical deformation within the thoracic cage. While investigating deformable slice-to-volume registration \citep{uus_deformable_2020} could yield further improvements, it introduces a significant computational penalty that currently conflicts with our goal of rapid clinical turnaround.

The \textbf{SPARC} pipeline is currently tailored for third-trimester fetuses where gross fetal mobility is relatively restricted. Recent developments combining slice-to-volume reconstruction with initial landmark-based registration have shown promise in handling the larger inter-slice motion characteristic of earlier gestational ages \citep{uus_automated_2022}, providing a clear pathway for future work to expand the applicable gestational age range.

\section{Conclusion}

We presented a hybrid pipeline combining physics-informed slice-to-volume reconstruction with deep learning components for automated 3D$+$time fetal cardiac cine reconstruction. By exploiting the temporal consistency afforded by Doppler gating, the proposed model achieves a tenfold reduction in computational complexity relative to existing approaches without loss in reconstruction quality and superior generalisation performance on held-out data. Deep learning components were trained under a restricted annotation budget via transfer learning from a larger source domain, enabling fully automatic processing in $82.6\%$ of clinical cases, reaching $100\%$ success for all cases meeting minimal inclusion criteria through efficient manual intervention, with a mean end-to-end processing time of $7.1 \pm 1.3$ min compatible with clinical deployment requirements. The current pipeline enables visualisation of intracardiac anatomy and function. Future work will focus on extracting downstream clinical metrics, such as ventricular volumes, ejection fraction, and flow velocity maps, to further establish its diagnostic value in assessing congenital heart disease.

The complete \textbf{SPARC} pipeline is publicly available as a Docker container\footnote{\href{https://hub.docker.com/r/aboutill/sparc}{https://hub.docker.com/r/aboutill/sparc}} and is currently deployed at our institution as a clinical diagnostic and research tool for fetal cardiac magnetic resonance imaging.

\section{Acknowledgement}
This work was funded by core funding from the Wellcome/EPSRC Centre for Medical Engineering [WT203148/Z/16/Z] and by the National Institute for Health and Care Research (NIHR) Clinical Research Facility (CRF) and HealthTech Research Centre in Cardiovascular and Respiratory Medicine (HRC) at Guy’s and St Thomas’ NHS Foundation Trust. The views expressed are those of the author(s) and not necessarily those of the NHS, the NIHR or the Department of Health and Social Care.

\bibliographystyle{cas-model2-names.bst}
\bibliography{references.bib}

\setcounter{table}{0}
\setcounter{section}{0}
\setcounter{figure}{0}
\setcounter{equation}{0}
\renewcommand{\thetable}{S\arabic{table}}
\renewcommand{\thesection}{S\arabic{section}}
\renewcommand{\thefigure}{S\arabic{figure}}
\renewcommand{\theequation}{S\arabic{equation}}

\section*{\Large{Supplementary Material}}

\begin{figure*}[h!]
\centering
\begin{adjustbox}{width=\columnwidth}
\begin{tikzpicture}[every node/.style={inner sep=0,outer sep=0}]

\begin{scope}[spy using outlines=
      {circle, magnification=1.25, size=.35cm, connect spies, rounded corners}]
      
\node[inner sep=0pt] at (0,0)
    {\includegraphics[width=.047\textwidth,angle=-90]{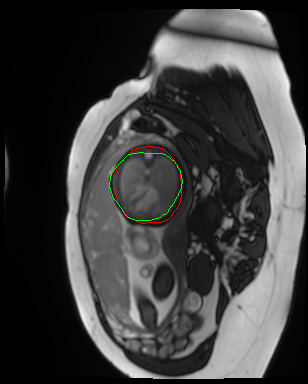}};
\node[inner sep=0pt] at (0,-.85)
    {\includegraphics[width=.047\textwidth,angle=-90]{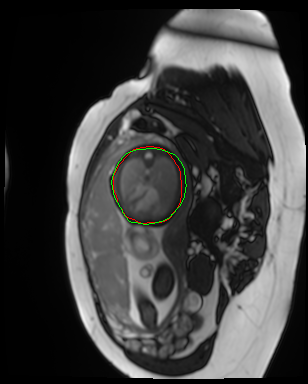}};
\node[inner sep=0pt] at (0,-1.7) 
    {\includegraphics[width=.047\textwidth,angle=-90]{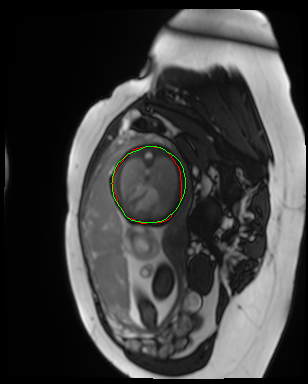}};
\node[inner sep=0pt] at (0,-2.55)
    {\includegraphics[width=.047\textwidth,angle=-90]{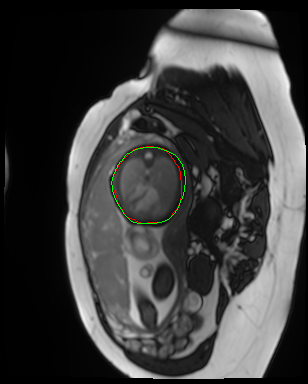}};

\spy [cyan] on (0.02,0.02) in node [left] at (-0.12,0.2);
\spy [cyan] on (0.02,-.83) in node [left] at (-0.12,-.65);
\spy [cyan] on (0.02,-1.68) in node [left] at (-0.12,-1.5);
\spy [cyan] on (0.02,-2.53) in node [left] at (-0.12,-2.35);

\end{scope}

\begin{scope}[spy using outlines=
      {circle, magnification=1.25, size=.3cm, connect spies, rounded corners}]
      
\node[inner sep=0pt] at (.85,0)
    {\includegraphics[width=.047\textwidth,angle=-90]{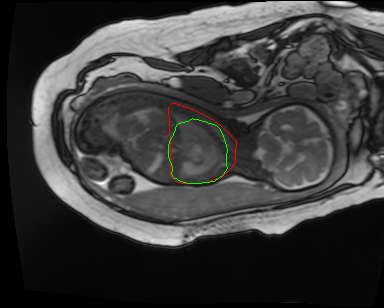}};
\node[inner sep=0pt] at (.85,-.85)
    {\includegraphics[width=.047\textwidth,angle=-90]{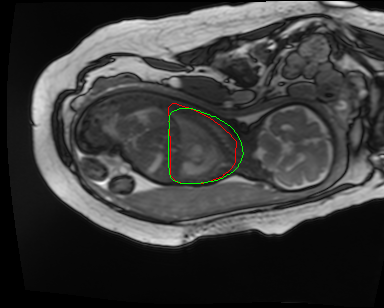}};
\node[inner sep=0pt] at (.85,-1.7) 
    {\includegraphics[width=.047\textwidth,angle=-90]{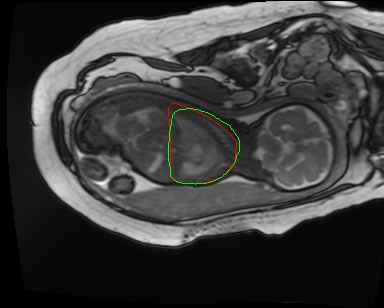}};
\node[inner sep=0pt] at (.85,-2.55)
    {\includegraphics[width=.047\textwidth,angle=-90]{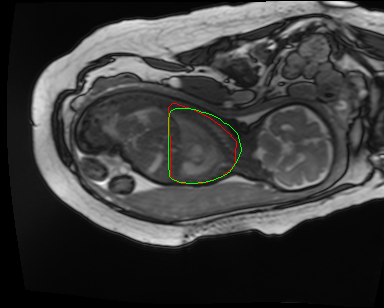}};

\spy [cyan] on (0.87,-0.005) in node [left] at (.84,0.23);
\spy [cyan] on (0.87,-0.855) in node [left] at (.84,-.62);
\spy [cyan] on (0.87,-1.705) in node [left] at (.84,-1.47);
\spy [cyan] on (0.87,-2.555) in node [left] at (.84,-2.32);

\end{scope}

\begin{scope}[spy using outlines=
      {circle, thick, magnification=1.1, size=.29cm, connect spies, rounded corners}]
      
\node[inner sep=0pt] at (1.5175,0)
    {\includegraphics[width=.047\textwidth,angle=-90]{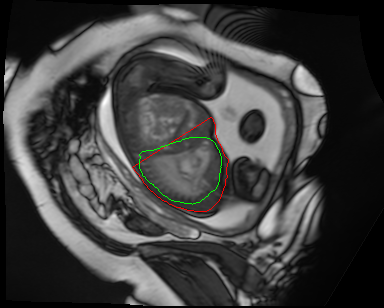}};
\node[inner sep=0pt] at (1.5175,-.85)
    {\includegraphics[width=.047\textwidth,angle=-90]{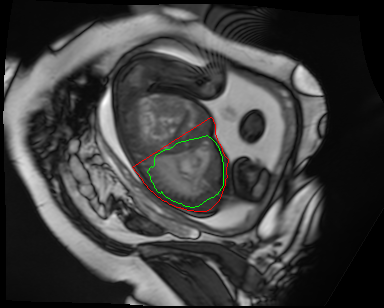}};
\node[inner sep=0pt] at (1.5175,-1.7) 
    {\includegraphics[width=.047\textwidth,angle=-90]{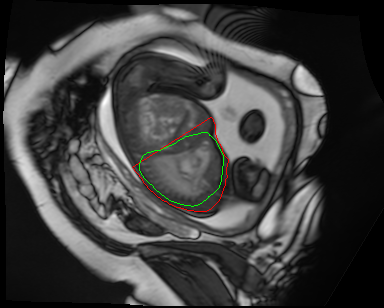}};
\node[inner sep=0pt] at (1.5175,-2.55)
    {\includegraphics[width=.047\textwidth,angle=-90]{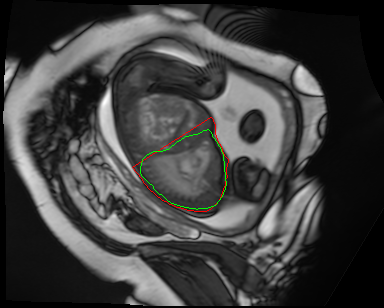}};

\spy [cyan] on (1.495,0.01) in node [left] at (1.82,-0.24);
\spy [cyan] on (1.495,-.84) in node [left] at (1.82,-1.09);
\spy [cyan] on (1.495,-1.69) in node [left] at (1.82,-1.94);
\spy [cyan] on (1.495,-2.54) in node [left] at (1.82,-2.79);

\end{scope}

\node[inner sep=0pt,rotate=90] at (-.6,0) {\scalebox{.35}{Source-only}};
\node[inner sep=0pt,rotate=90] at (-.6,-.85) {\scalebox{.35}{Target-only}};
\node[inner sep=0pt,rotate=90] at (-.6,-1.7) {\scalebox{.35}{Joint}};
\node[inner sep=0pt,rotate=90] at (-.6,-2.55) {\scalebox{.35}{Transfer}};

\node[inner sep=0pt] at (0,.475) {\scalebox{.35}{Case 1}};
\node[inner sep=0pt] at (.85,.475) {\scalebox{.35}{Case 2}};
\node[inner sep=0pt] at (1.5175,.475) {\scalebox{.35}{Case 3}};

\end{tikzpicture}
\end{adjustbox}
\caption{Qualitative comparison of thoracic segmentation contours for three target domain subjects across the four training strategies (source-only, target-only, joint, and transfer). Time-averaged stacks were selected to represent distinct stack orientations and fetal position. Ground truth delineations are in red (\textcolor{red} {\raisebox{1.8pt}{\rule{5pt}{1pt}}}) while predictions appear in green (\textcolor{green}{\raisebox{1.8pt}{\rule{5pt}{1pt}}}).}
\label{fig:thorax_seg}
\end{figure*}

\begin{figure*}[h!]
\centering
\begin{adjustbox}{width=\textwidth}
\begin{tikzpicture}[every node/.style={inner sep=0,outer sep=0}]

\node[inner sep=0pt] at (0,0)
    {\includegraphics{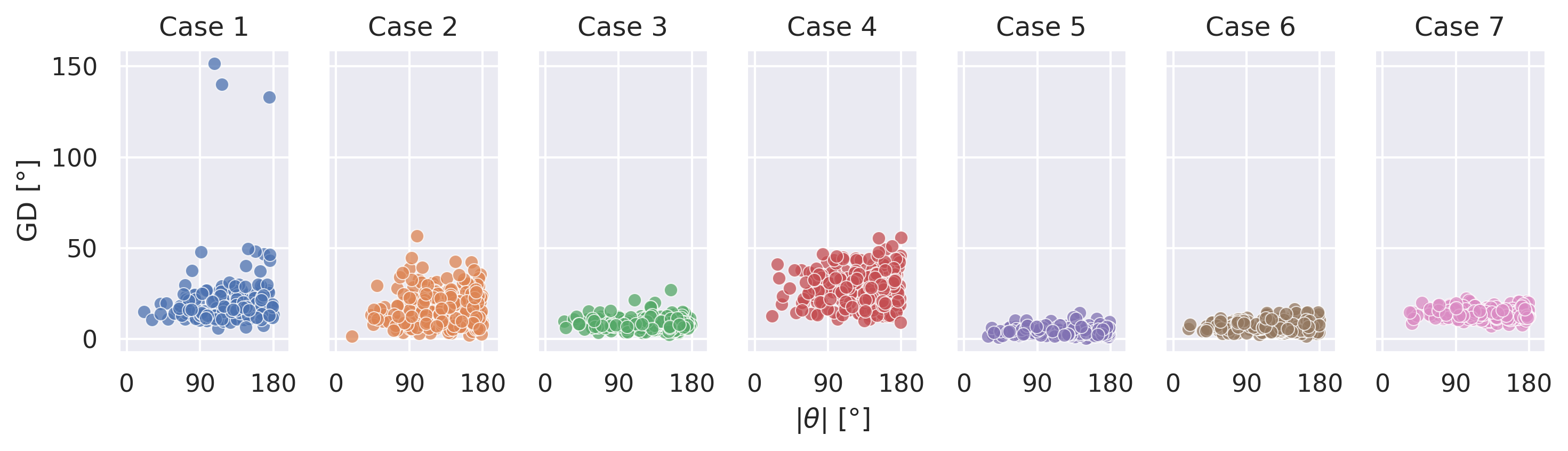}};

\end{tikzpicture}
\end{adjustbox}
\caption{Geodesic distance (GD) of the Transfer approach as a function of simulated rotation magnitude $|\theta|$, evaluated on the held-out target domain test set (Doppler-gated; $n = 7$ subjects, 250 randomly sampled rigid transformations per subject). Each point represents one simulated transformation, with subjects colour-coded.}
\label{fig:reorientation_subject}
\end{figure*}

\begin{figure*}[h!]
\centering
\begin{adjustbox}{width=\columnwidth}
\begin{tikzpicture}[every node/.style={inner sep=0,outer sep=0}]

\node[inner sep=0pt] at (0,0)
    {\includegraphics[width=.05\textwidth]{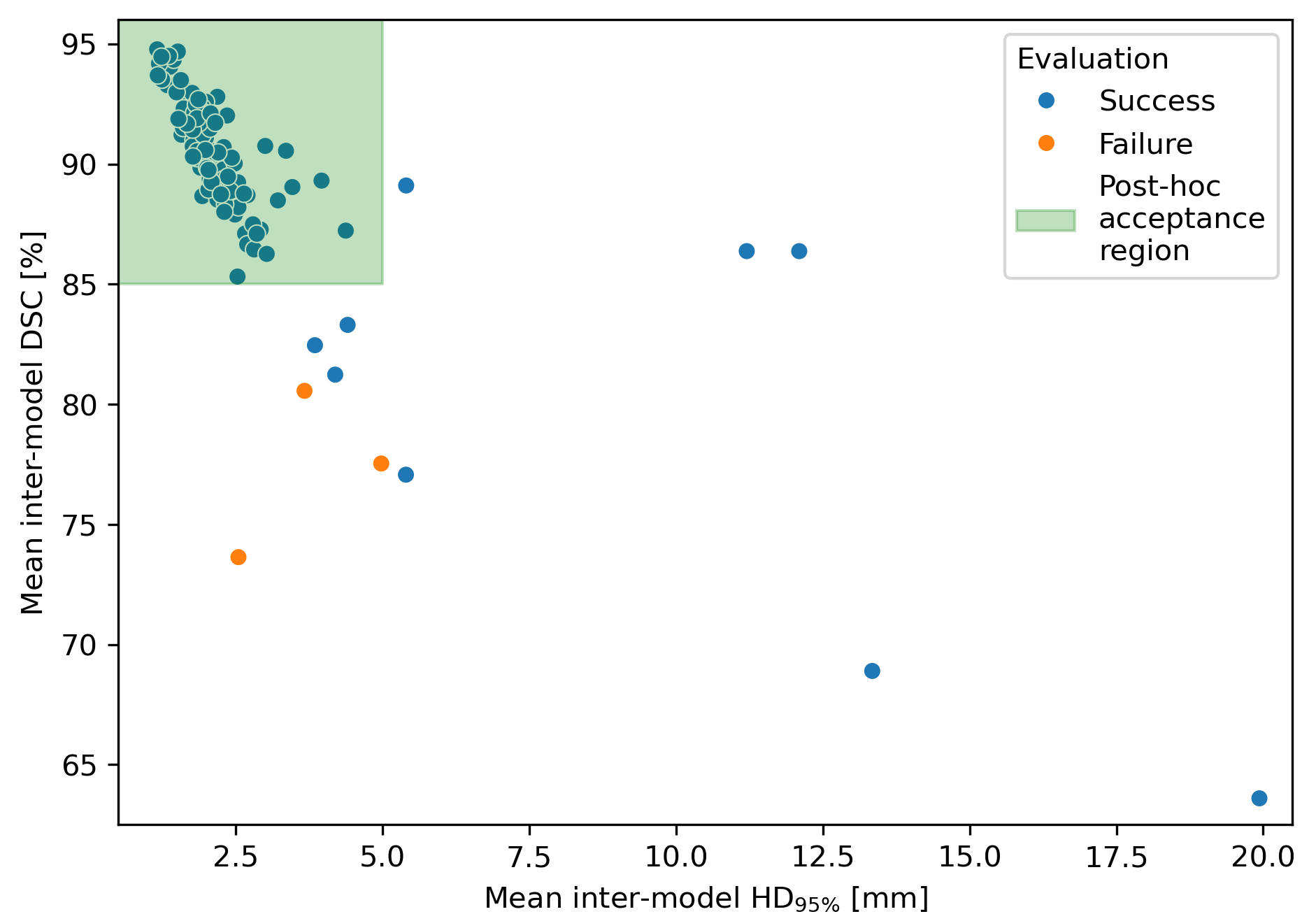}};

\node[inner sep=0pt] at (0,-.65)
    {\includegraphics[width=.05\textwidth]{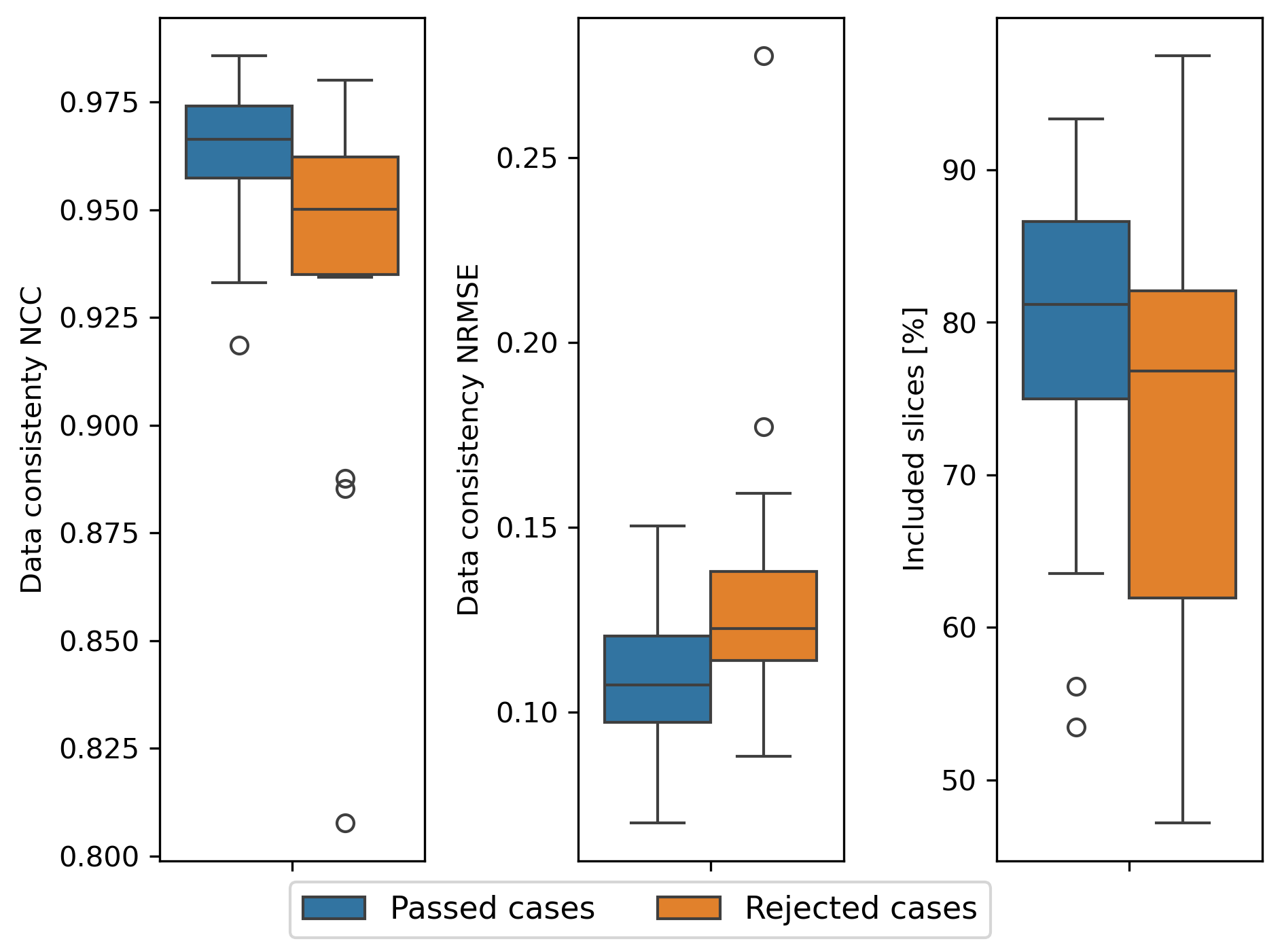}};

\node[inner sep=0pt] at (0,-1.3)
    {\includegraphics[width=.05\textwidth]{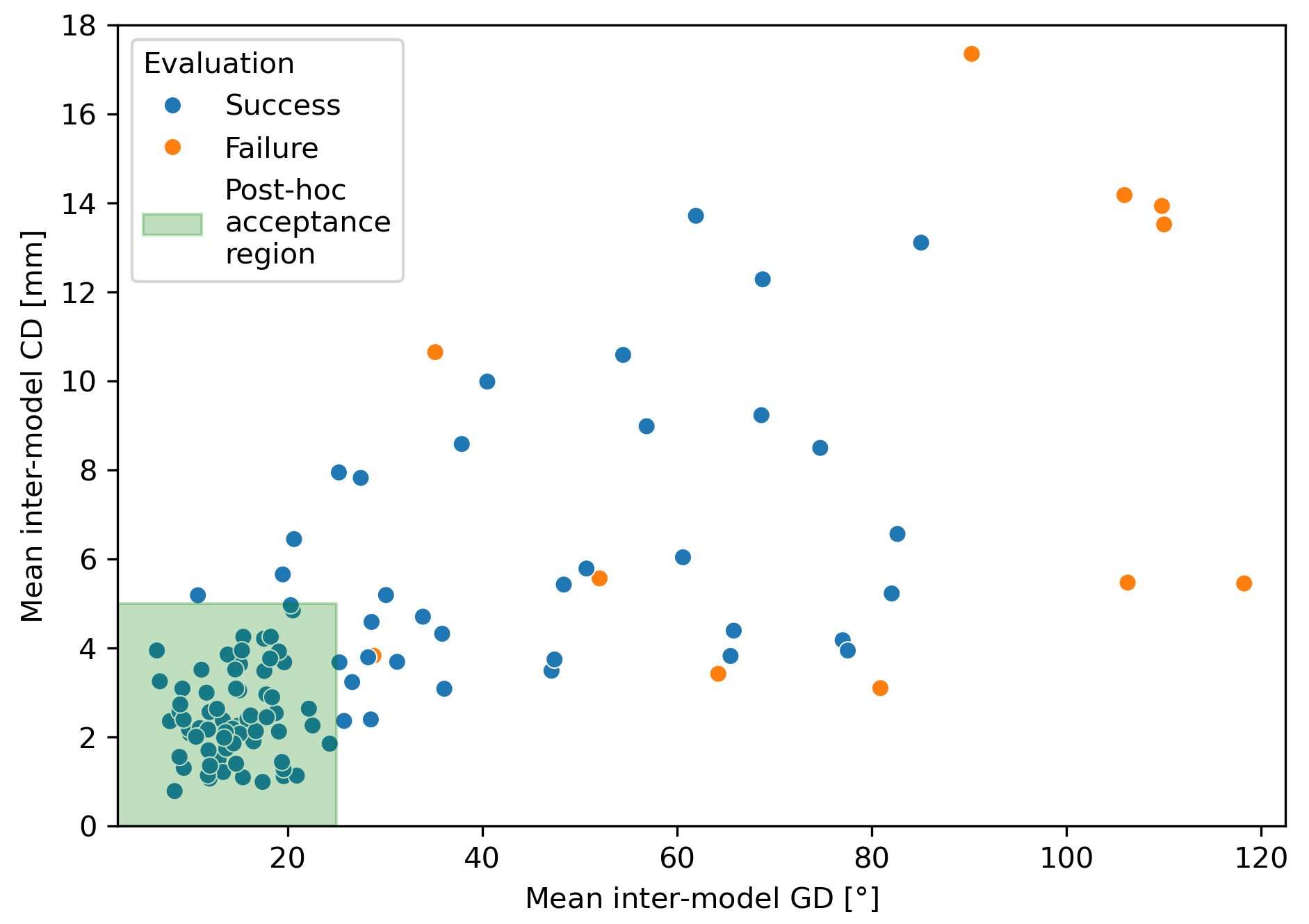}};

\end{tikzpicture}
\end{adjustbox}
\caption{Automatic quality control metrics for the three main \textbf{SPARC} pipeline stages, evaluated on the end-to-end clinical cohort ($n = 121$ subjects). \textbf{A)} Thoracic segmentation: mean pairwise inter-model DSC versus $\mathrm{HD}_{95\%}$ for each subject, colour-coded by segmentation outcome. The post-hoc acceptance region is define by DSC $> 85\%$ and $\mathrm{HD}_{95\%} < 5$\,mm. \textbf{B)} Cine reconstruction: data consistency metrics (NCC, NRMSE) and proportion of included slices for pipeline-accepted ($n = 121$) and pipeline-rejected ($n = 18$) cases. \textbf{C)} Reorientation: mean pairwise inter-model GD versus CD for each subject, colour-coded by reorientation outcome. The post-hoc acceptance region was defined as GD $< 25\degree$ and CD $< 5$\,mm.}
\label{fig:pipeline_qc}
\end{figure*}

\end{document}